\documentclass[]{bytedance_seed}

\usepackage[toc,page,header]{appendix}

\usepackage{minitoc}
\usepackage{type1cm}
\RequirePackage[T1]{fontenc}
\usepackage{amssymb}
\usepackage{pifont} 
\usepackage{algorithm} 
\usepackage{algpseudocode} 
\usepackage{makecell}
\usepackage[table]{xcolor}
\usepackage{wrapfig}
\usepackage{multirow}
\usepackage{colortbl}
\usepackage{makecell}
\usepackage{soul}

\newcommand{\cmark}{\ding{51}} 
\newcommand{\xmark}{\ding{55}} 

\newcommand{\sref}[1]{\S\ref{#1}}

\definecolor{lightgray}{HTML}{E8E8E8}
\definecolor{lightgreen}{HTML}{D2F3D7}
\definecolor{lightred}{HTML}{F9D1D4}
\definecolor{arm1}{HTML}{C0392B}  
\definecolor{arm2}{HTML}{C77700}  
\definecolor{arm3}{HTML}{1A7F37}  
\usepackage{listings}
\usepackage{color}

\usepackage{mdframed}
\definecolor{mdproofbg}{rgb}{0.95,0.95,0.95}
{\begin{mdframed}[backgroundcolor=mdproofbg,linewidth=0]\begin{proof}}%
{\end{proof}\end{mdframed}}

\definecolor{mdworkingbg}{rgb}{1.0,0.95,0.95}
{\begin{mdframed}[backgroundcolor=mdworkingbg,linewidth=0]\begin{minipage}{\columnwidth}}%
{\end{minipage}\end{mdframed}}

\usepackage{xpatch}
\makeatletter
\AtBeginDocument{\xpatchcmd{\@thm}{\thm@headpunct{.}}{\thm@headpunct{:}}{}{}} 
\AtBeginDocument{\xpatchcmd{\@thm}{\fontseries\mddefault\upshape}{}{}{}} 
\makeatother

\definecolor{bgCode}{rgb}{0.98, 0.98, 0.98}
\definecolor{codegray}{rgb}{0.5,0.5,0.5}
\lstnewenvironment{pycode}%
{  \noindent
	\minipage{\linewidth} 
	\bigskip 
	\lstset{language=Python,numbers=left,numbersep=5pt}
	\raggedright
	\scriptsize{\textsf{Python Code}\vspace{-1.75ex}}
	\raggedleft}%
{\smallskip \endminipage}

\title{GenFirst: Generation Before Reconstruction for \\Stable End-to-End Latent Generative Modeling}
\author[1,2]{Guangting Zheng}
\author[2]{Yiyuan Zhang}
\author[2]{Tao Yang}
\author[2]{Yunpeng Chen}
\author[2,\dagger]{Rui Zhu}
\author[1]{\\Jiajun Deng}
\author[1,\ddag]{Yanyong Zhang}
\affiliation[1]{University of Science and Technology of China}
\affiliation[2]{ByteDance Seed}

\contribution[\dagger]{Project lead}
\contribution[\ddag]{Corresponding author}

\abstract{
Latent generative models typically follow a two-stage pipeline, training a variational autoencoder for reconstruction and then a generative model on the frozen latent space. 
Since reconstruction-optimized latents are not necessarily generation-friendly, jointly training both models is an appealing alternative. 
However, direct end-to-end training remains challenging, as it is prone to \textbf{latent collapse} and faces an \textbf{generation--reconstruction conflict}. 
In this work, we revisit this problem by analyzing how different objectives shape the latent space and identify two key insights. 
First, the entropy term in the Kullback-Leibler divergence (KL) objective is essential for preventing collapse: reconstruction and prior fitting both tend to shrink the posterior, while entropy preserves non-degenerate latent uncertainty.
Second, reconstruction and generation exhibit asymmetric learning dynamics: reconstruction is fast and strongly supervised, whereas generation is slower and harder to optimize. 
Motivated by these insights, we achieve the first direct end-to-end training without latent collapse and further propose GenFirst, a simple generation-before-reconstruction strategy. 
The generative objective first shapes the latent space under weak reconstruction pressure, after which reconstruction is progressively strengthened to recover visual details. 
We validate this strategy using both continuous autoregressive priors with exact likelihoods and SiT priors with implicit likelihoods. 
With our end-to-end objective and GenFirst, SiT achieves a gFID of 0.97 with CFG and 1.45 without CFG on ImageNet $256\times256$, while MMDiT reaches a GenEval score of 0.90 on text-to-image generation.
Beyond image generation, we further extend the framework to shared visual latents for generation and representation learning, and to continuous unified text--image generation.
These results demonstrate the generality of stable end-to-end latent learning across generative priors and modalities.
}

\date{\today}
\correspondence{\email{zgt@mail.ustc.edu.cn}}

\checkdata[Project Page]{\url{https://tom-zgt.github.io/GenFirst/}}

\begin{document}
\maketitle

\section{Introduction}\label{sec:intro}
\begin{figure}
    \centering
    \includegraphics[width=\linewidth]{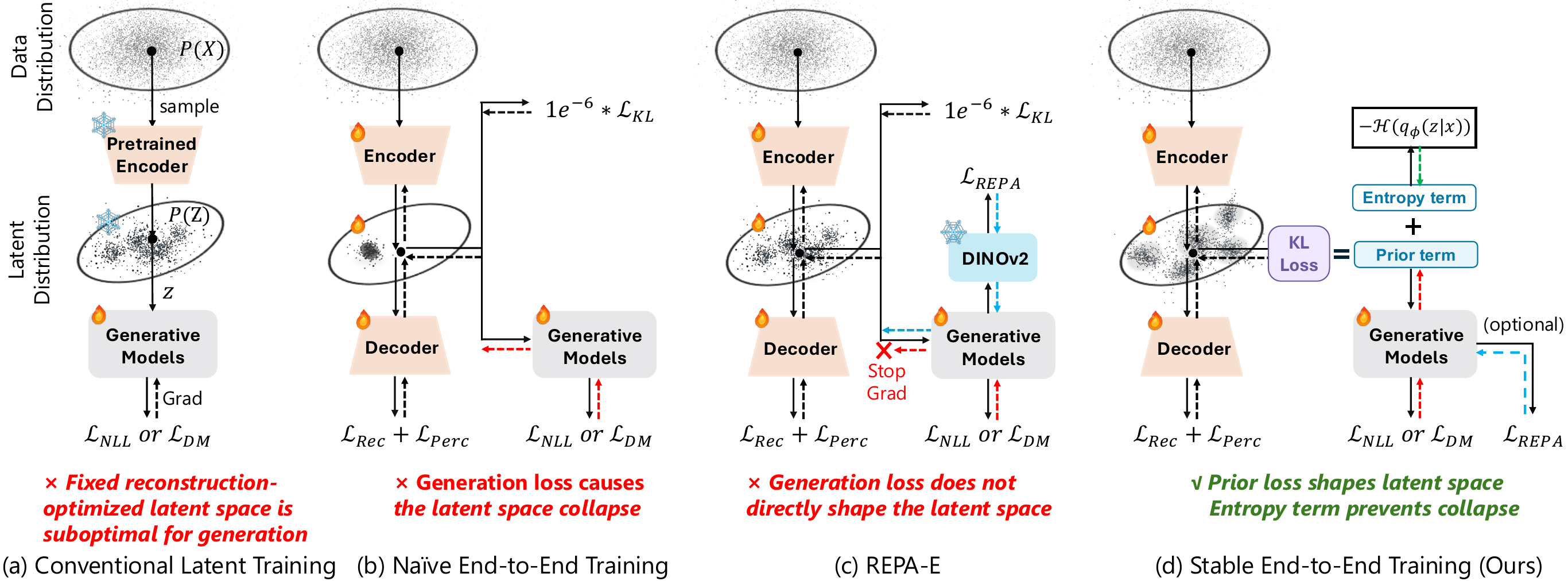}
    \caption{\textbf{Comparison of latent generative modeling paradigms.}
    (a) Conventional two-stage methods train generative models on a fixed, reconstruction-optimized latent space that may be suboptimal for generation.
    (b) Naïve end-to-end training allows the generative loss to update the encoder but causes latent collapse.
    (c) REPA-E avoids collapse by stopping generative gradients and reshaping the latent space through an external representation-alignment loss, preventing the generative objective from directly shaping the latent space.
    (d) Our method enables direct end-to-end training: the prior loss shapes a generation-friendly latent space, while the entropy term prevents collapse.}
    \label{fig:intro}
\end{figure}
Latent generative models~\cite{ldm,maskdit,uvit,dit,sit,decoupled_dit,mar,llamagen,var,starflow} have become a dominant paradigm for high-resolution image generation. 
Instead of modeling images directly in pixel space, they first use a variational autoencoder (VAE)~\cite{vae} or visual tokenizer to compress raw pixels into a low-dimensional latent space and then train a generative model on this latent distribution. Such a two-stage pipeline greatly reduces modeling complexity and has enabled huge progress in image generation. 
However, recent works~\cite{repa,vavae,repa_e,simflow,rae} show that a VAE latent space optimized for reconstruction is not necessarily optimal for generation. 
Reconstruction tends to preserve instance-level details, making the latent distribution overly dispersed and harder for generative models to fit and sample from. 
This mismatch between reconstruction and generation has motivated diverse approaches, including improved autoencoders~\cite{vavae}, representation alignment~\cite{repa}, pixel-space generation~\cite{zheng2025farmer,jit,pixeldit,deco,cai2026hidream}, and end-to-end training~\cite{repa_e,simflow,chu2026end}. 
Despite their different designs, these approaches all address a shared question: what representation is best suited for generative modeling?

A natural solution is to jointly train the VAE and generative model, allowing the generative objective to directly shape the latent space.
JetFormer~\cite{tschannen2024jetformer} and FARMER~\cite{zheng2025farmer} use invertible normalizing flows~\cite{nvp,tarflow} instead of VAEs, while SimFlow~\cite{simflow} jointly trains a VAE and normalizing flow with fixed VAE variance.
REPA-E~\cite{repa_e} jointly trains diffusion models with a VAE, but observes severe latent collapse when gradients from the generative model are propagated to the VAE. 
It therefore stops the diffusion gradient and instead uses representation alignment for indirect supervision. 
Together, these designs highlight the promise of end-to-end latent generative modeling, while leaving its feasibility and optimization mechanism unclear.

In this paper, we therefore ask the first fundamental question: \textbf{Is direct end-to-end training of VAEs and generative models feasible?}
REPA-E~\cite{repa_e} observed that naively propagating the generative loss through the VAE causes latent collapse.
To answer this question, we revisit end-to-end training from the perspective of the VAE objective and identify a previously overlooked prior--entropy imbalance (\sref{subsec:prior_entropy}).
LDM~\cite{ldm} established a practical VAE training recipe widely adopted by modern latent generative models, where a small KL weight weakly regularizes the latent space.
The KL regularizer in the VAE objective consists of a prior-fitting term and a negative posterior-entropy term.
The former encourages latent samples to match the Gaussian prior, whereas the latter preserves posterior uncertainty and prevents degeneracy.
In naive end-to-end training, the added generative objective introduces a strong prior-fitting force, while posterior entropy remains suppressed by the inherited small KL weight.
This imbalance drives the posterior variance toward zero and the latent representations toward a nearly constant value, ultimately causing latent collapse (\sref{subsec:analysis}). 
Our analysis turns latent collapse from an empirical failure mode into a diagnosable and correctable imbalance in the objective: stable end-to-end training requires explicitly preserving posterior entropy and balancing it against prior fitting.

\begin{figure}[t]
    \centering
    \includegraphics[width=\linewidth]{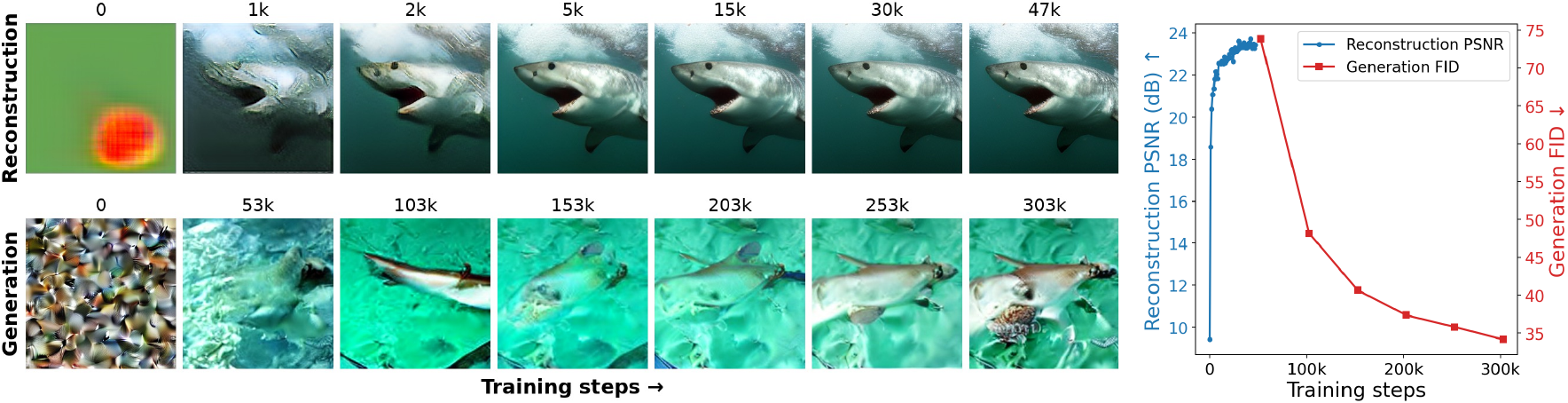}
    \caption{
     \textbf{Visualization of the asymmetric learning dynamics between reconstruction and generation.} 
     Reconstruction reaches good image fidelity within much fewer training steps, while generation requires a longer optimization process to learn a sampleable latent distribution.
    }
    \label{fig:rg_visual}
\end{figure}

After resolving the collapse issue, end-to-end training still faces a generation--reconstruction conflict~\cite{vavae}: reconstruction favors preserving instance-specific details, whereas generation favors a smoother and more modelable latent distribution, potentially discarding information needed for faithful reconstruction. 
This results in a generation--reconstruction trade-off, where improving one objective often degrades the other (\sref{subsec:no_single_optimum}).

Therefore, we ask a second fundamental question: \textbf{How can we mitigate the generation--reconstruction conflict to learn a latent space better suited for generative modeling?}
As shown in \cref{fig:rg_visual}, our experiments reveal asymmetric learning dynamics between reconstruction and generation (\sref{subsec:loss_balancing}).
Reconstruction improves rapidly under direct supervision, whereas generation requires longer optimization to learn a globally sampleable latent distribution.
Strong reconstruction pressure early in training can therefore anchor the encoder to a reconstruction-oriented latent space that is difficult for the generative model to learn.
Conversely, maintaining excessive generative pressure can limit the recovery of visual details needed for high-quality decoding.
These observations suggest that the key is not only how the two objectives are weighted, but also when each objective shapes the latent space.

This observation motivates \textbf{GenFirst}, a simple generation-before-reconstruction strategy for our end-to-end training framework.
Rather than first learning a reconstruction-oriented latent space and then forcing the generative model to chase it, GenFirst allows the generative objective to shape the latent distribution early, under weak reconstruction pressure and explicit entropy preservation. 
Once the latent distribution can be effectively modeled by the generative prior, reconstruction is progressively strengthened to recover visual details.
This ordering aligns with the asymmetric learning dynamics of the two objectives: generation needs to shape the latent geometry early, while reconstruction, being faster and more directly
supervised, can be strengthened later.

We instantiate GenFirst in two representative settings. 
First, we adopt a continuous autoregressive prior from FARMER~\cite{zheng2025farmer}, where exact likelihood is available and the
prior--entropy balance can be directly analyzed.
Second, we apply GenFirst to SiT~\cite{sit}, a flow-matching prior with an implicit likelihood, to evaluate whether the strategy generalizes beyond exact-likelihood models. 
Across class-to-image and text-to-image generation, GenFirst consistently improves the generation--reconstruction trade-off, providing a practical and reliable recipe for stable end-to-end latent generative modeling.
Beyond image generation, we further extend our end-to-end formulation to shared visual latents that support generation, representation learning, and reconstruction.
We also apply it to continuous unified text--image modeling, demonstrating its generality across modalities.

We summarize our contributions as follows:
\begin{itemize}
    \item We identify a prior--entropy imbalance as the key cause of latent collapse in direct end-to-end training and show that preserving posterior entropy is essential for preventing collapse.
    \item 
    We reveal the asymmetric learning dynamics of generation and reconstruction and 
    propose a GenFirst strategy that first shapes a generation-friendly latent space and then progressively recovers visual details.
    \item We validate GenFirst with both exact-likelihood continuous autoregressive priors and SiT-based flow-matching priors, achieving a FID of 0.97 on ImageNet and a GenEval score of 0.90 on text-to-image generation. We also validate its applicability across objectives and modalities by further experiments.
\end{itemize}

\section{Related Work}

\textbf{Latent Generative Models.}
Latent generative models~\cite{ldm,uvit,maskdit,dit,sit,llamagen,mar,var,himar,starflow} use image tokenizers, such as VAEs~\cite{ldm,vavae}, to compress images into a lower-dimensional latent space, which simplifies generative modeling while leaving high-frequency details to the tokenizer. 
However, recent studies~\cite{repa,vavae,repa_e} show that a VAE latent space optimized for reconstruction is not necessarily optimal for generation.

\textbf{Representation Learning for Generation.}
Recent works explore how to obtain more generation-friendly representations.
REPA~\cite{repa} improves diffusion models by aligning their intermediate hidden states with features from pretrained visual encoders such as DINOv2~\cite{dino}. 
VA-VAE~\cite{vavae} applies representation alignment to VAE training, while REPA-E~\cite{repa_e} further propagates the REPA loss through the diffusion model to align the VAE latent space in an end-to-end manner. 
Another line of work~\cite{tarflow,wang2025pixnerd,zheng2025farmer,jit,pixeldit,deco} revisits the original pixel representation and removes the VAE entirely. 
However, pixel-space generative models still face the challenge of modeling high-dimensional data and often remain less efficient than latent generative models.

\textbf{End-to-End Generative Modeling.}
Several works study joint training of autoencoders and generative models. 
LSGM~\cite{lsgm} jointly trains a VAE with score-based generative models using a complex design. 
REPA-E~\cite{repa_e} jointly trains a VAE and a diffusion model by stopping the diffusion gradient to the VAE and introducing a representation alignment loss.
JetFormer~\cite{tschannen2024jetformer} and FARMER~\cite{zheng2025farmer} jointly train lossless normalizing flows~\cite{kolesnikov2024jet,tarflow} and autoregressive models~\cite{llamagen} to map the data distribution to a latent distribution modeled by AR priors. 
SimFlow and MIMFlow~\cite{simflow,chen2026mimflow} jointly train a VAE and normalizing flows by fixing the VAE variance. 
Unified Latents \cite{heek2026unified} first jointly learns bitrate-controlled latents with an encoder, diffusion prior, and decoder, then freezes the encoder and retrains a larger generation-oriented latent diffusion prior for efficient, high-quality synthesis.
However, existing works still lack a systematic understanding of direct end-to-end training between VAEs and generative models, especially the cause of latent collapse, the conflict between generation and reconstruction in shaping the latent space, and a simple general training strategy. 
In this paper, we revisit end-to-end latent generative modeling from the VAE objective, identify a prior-entropy imbalance as the key cause of collapse, and propose a generation-before-reconstruction strategy.

\section{Revisiting End-to-End Latent Generative Modeling}
\label{sec:revisiting_e2e}
A standard latent generative model is trained in two stages. 
In the first stage, a VAE is trained to encode images into a continuous latent space. 
Given an image $x$, the encoder $q_\phi(z|x)$ predicts a mean $\mu$ and a variance $\sigma^2$, which define a Gaussian posterior $\mathcal{N}(\mu, \mathrm{diag}(\sigma^2))$. 
A latent vector $z$ is then sampled from this posterior and decoded by $D_\psi$ to reconstruct the image $\hat{x}=D_\psi(z)$. 
The VAE is optimized with a reconstruction loss, typically an $\ell_1$ loss between $x$ and $\hat{x}$, together with perceptual losses computed on $\hat{x}$, including LPIPS and GAN losses. 
In practice, a weak KL regularization term, usually with a very small weight such as $10^{-6}$~\cite{ldm}, is also applied to $\mu$ and $\sigma^2$ to mildly constrain the posterior distribution.
The VAE training objective is
\begin{equation}
    \mathcal{L}_{\mathrm{VAE}}
    =
    \lambda_{\mathrm{rec}}\mathcal{L}_{\mathrm{rec}}
    +
    \lambda_{\mathrm{LPIPS}}\mathcal{L}_{\mathrm{LPIPS}}
    +
    \lambda_{\mathrm{GAN}}\mathcal{L}_{\mathrm{GAN}}
    +
    \lambda_{\mathrm{KL}}
    D_{\mathrm{KL}}(q_\phi(z|x)\|p_0(z)).
\end{equation}
In the second stage, the pretrained VAE is frozen and a generative prior is trained on latents sampled from the encoder posterior.
For an exact-likelihood generative prior $p_\theta(z)$, the prior loss is
\begin{equation}
    \mathcal{L}_{\mathrm{prior}}
    =
    \mathbb{E}_{x,\,z\sim q_\phi(z|x)}
    \left[-\log p_\theta(z)\right].
\end{equation}
A natural end-to-end alternative is to jointly optimize the VAE and the prior:
\begin{equation}
    \mathcal{L}_{\mathrm{naive}}
    =
    \mathcal{L}_{\mathrm{VAE}}
    +
    \lambda_{\mathrm{prior}}\mathcal{L}_{\mathrm{prior}}.
    \label{eq:naive_e2e}
\end{equation}
While this lets the prior directly shape the latent space, it also changes the forces acting on the encoder posterior. 
Without proper balancing, the posterior can degenerate and cause latent collapse.

\subsection{Prior-Entropy Imbalance Causes Latent Collapse}
\label{subsec:prior_entropy}
To understand why naive end-to-end training collapses, we revisit the KL term.
For the standard Gaussian prior $p_0(z)=\mathcal{N}(0,I)$, the KL divergence can be decomposed as
\begin{align}
    D_{\mathrm{KL}}(q_\phi(z|x)\|p_0(z))
    &=
    \mathbb{E}_{q_\phi}
    [\log q_\phi(z|x)-\log p_0(z)] \\
    &=
    \underbrace{
    \mathbb{E}_{q_\phi}[-\log p_0(z)]
    }_{\text{prior fitting}}
    -
    \underbrace{
    \mathcal{H}(q_\phi(z|x))
    }_{\text{entropy}}.
\end{align}
The prior-fitting term pulls latent samples toward high-probability regions of the reference prior $p_0(z)$.
The entropy term counteracts this effect by keeping the posterior from degenerating into point masses.

As described above, naive end-to-end training treats the VAE KL loss as an engineering regularizer and keeps the same tiny weight following LDM~\cite{ldm}, typically $10^{-6}$. 
This overlooks the entropy-preserving role of the KL term when a learned prior loss is introduced.
Once the generative objective is added, it introduces additional prior-fitting pressure and breaks the balance between prior fitting and posterior entropy: $\lambda_{\mathrm{prior}}+\lambda_{\mathrm{KL}}\gg\lambda_{\mathrm{KL}}$. 
This prior--entropy imbalance drives the posterior variance toward zero and pushes the posterior means to collapse to nearly the same code, resulting in a collapsed latent space.
Based on this analysis, we explicitly separate prior fitting and posterior entropy in \cref{eq:naive_e2e}, and use $\mathcal{L}_{\mathrm{prior}}$ to denote the generative objective that shapes the latent distribution:
\begin{equation}
    \mathcal{L}_{\mathrm{E2E}}
    =
    \lambda_{\mathrm{rec}}\mathcal{L}_{\mathrm{rec}}
    +
    \lambda_{\mathrm{LPIPS}}\mathcal{L}_{\mathrm{LPIPS}}
    +
    \lambda_{\mathrm{GAN}}\mathcal{L}_{\mathrm{GAN}}
    +
    \lambda_{\mathrm{prior}}\mathcal{L}_{\mathrm{prior}}
    -
    \lambda_{\mathrm{ent}}\mathcal{H}(q_\phi(z|x)).
    \label{eq:e2e_objective}
\end{equation}
\textbf{In this view, the entropy term is not a minor regularizer, but rather provides a counteracting force to learned prior fitting and prevents latent collapse.}

\begin{figure}
    \centering
    \includegraphics[width=\linewidth]{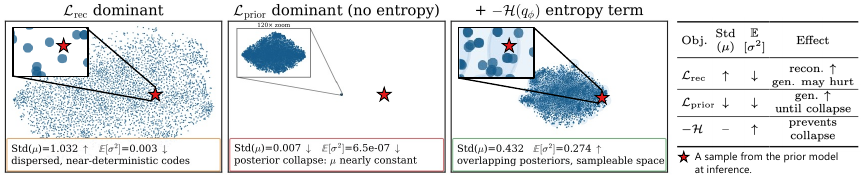}
    \caption{Qualitative effects of each objective on the latent posterior. Here $z=\mu_\phi(x)+\sigma_\phi(x)\epsilon$.}
    \label{fig:obj_effect}
\end{figure}

To further illustrate how different losses affect the latent space, we visualize two statistics: $\mathrm{Std}(\mu)$ and $\mathbb{E}[\sigma^2]$.
$\mathrm{Std}(\mu)$ measures how widely the posterior means are distributed across images, while $\mathbb{E}[\sigma^2]$ measures the average posterior uncertainty.
As shown in \cref{fig:obj_effect}, different objectives push the latent posterior in different directions.
When reconstruction dominates, $\mathrm{Std}(\mu)$ increases but $\mathbb{E}[\sigma^2]$ decreases: images become easier to reconstruct, but the latent space contains gaps between codes and may be difficult to sample from.
When prior fitting dominates without sufficient entropy, both $\mathrm{Std}(\mu)$ and $\mathbb{E}[\sigma^2]$ decrease, causing the posterior to collapse toward a nearly constant code.
Adding the entropy term increases $\mathbb{E}[\sigma^2]$, increases posterior overlap and reduces gaps between sampled codes, which prevents collapse and improves sampleability.
Additionally, we discuss a subtle shortcut solution behind the collapse behavior in \cref{app:collapse_shortcut}.

\subsection{Generation--Reconstruction Trade-off}
\label{subsec:no_single_optimum}
After preventing latent collapse with entropy, end-to-end training still faces a generation--reconstruction conflict.
To study this conflict, we keep the reconstruction-related loss weights the same as in the VAE training setting and vary only the prior loss weight.
As shown in \cref{fig:abl_1}(a), increasing $\lambda_{\mathrm{prior}}$ improves generation but hurts reconstruction: gFID improves from $120.60$ to $8.06$, while rFID worsens from $0.53$ to $4.20$.
This suggests that generation and reconstruction favor different operating points along the observed trade-off and cannot generally reach their individual optima at one prior weight.
We analyze this problem using the continuous autoregressive prior from FARMER~\cite{zheng2025farmer}, where the exact likelihood is available: $\mathcal{L}_{\mathrm{prior}}^{\mathrm{AR}}
    =
    \mathbb{E}_{z\sim q_\phi(z|x)}
    \left[
    -\sum_i \log p_\theta(z_i|z_{<i})
    \right]$.
This setting allows us to directly control the strength of the generative objective.
We then explore three natural strategies to mitigate this conflict, but find that they still fail to resolve it.

\begin{figure}[t]
    \centering

    \begin{subfigure}[t]{0.32\linewidth}
        \centering
        \includegraphics[width=\linewidth]{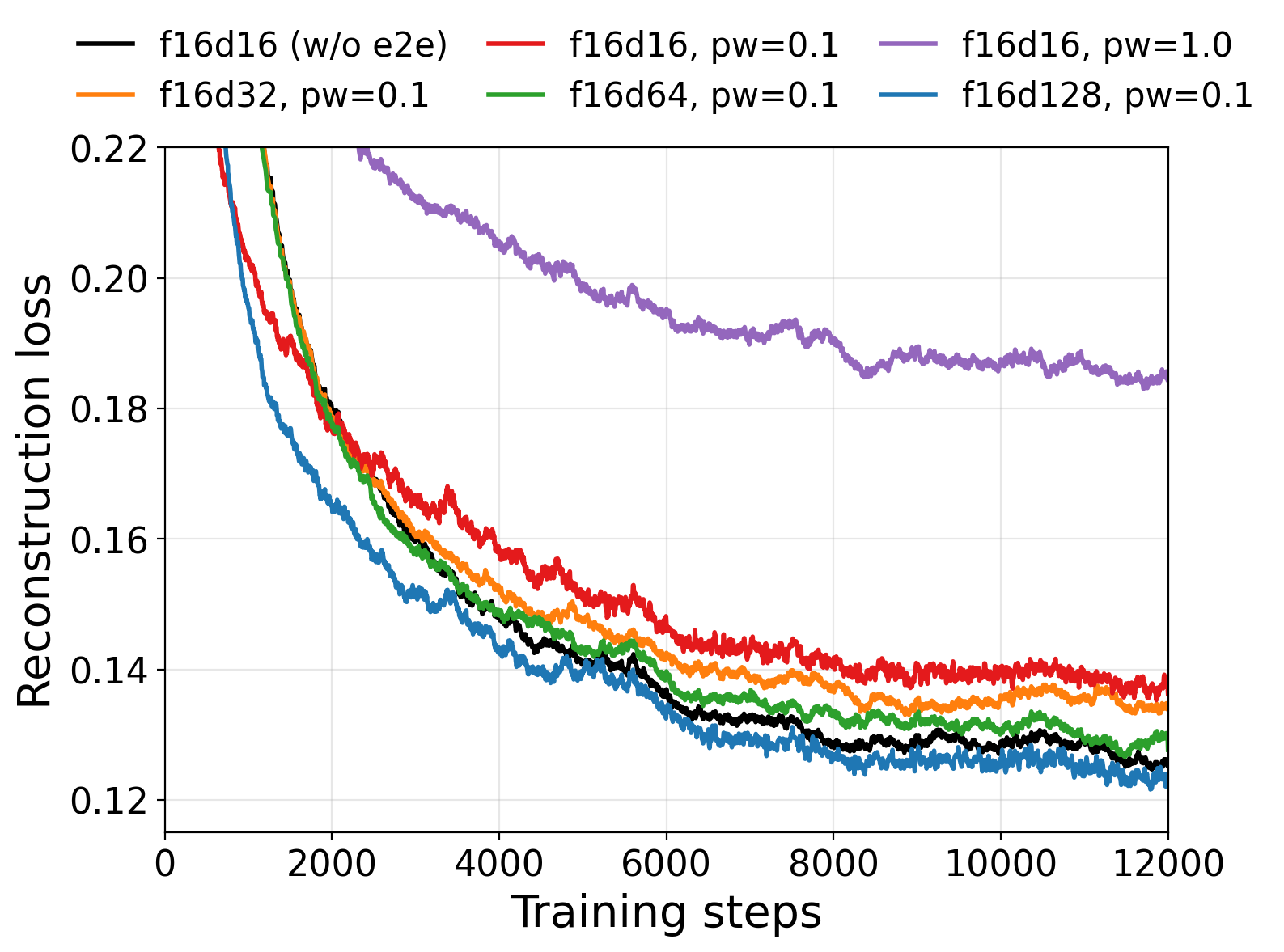}
        \caption{Latent dim}
        \label{fig:latent_dim}
    \end{subfigure}
    \hfill
    \begin{subfigure}[t]{0.32\linewidth}
        \centering
        \includegraphics[width=\linewidth]{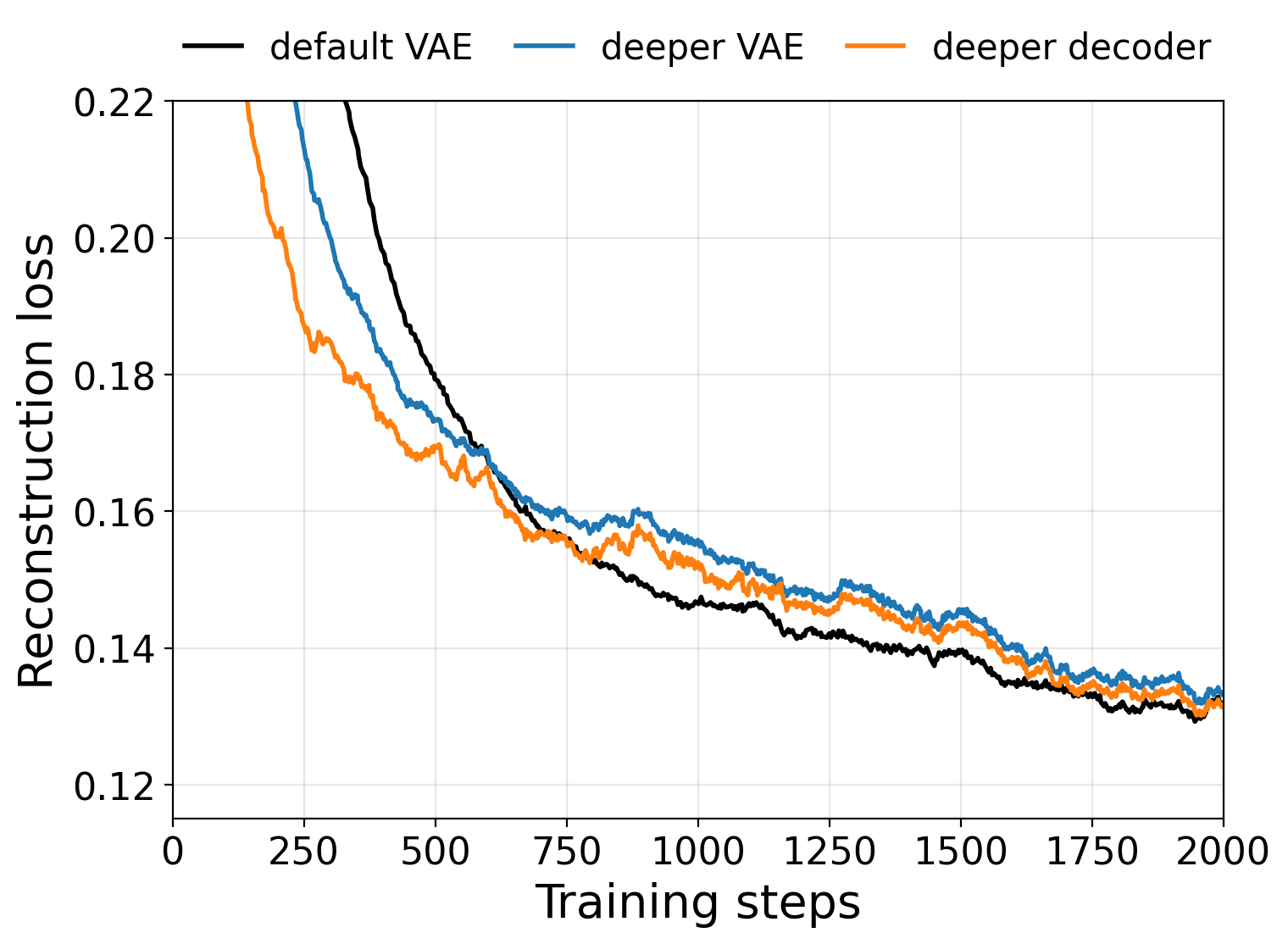}
        \caption{VAE size}
        \label{fig:vae_size}
    \end{subfigure}
    \hfill
    \begin{subfigure}[t]{0.32\linewidth}
        \centering
        \includegraphics[width=\linewidth]{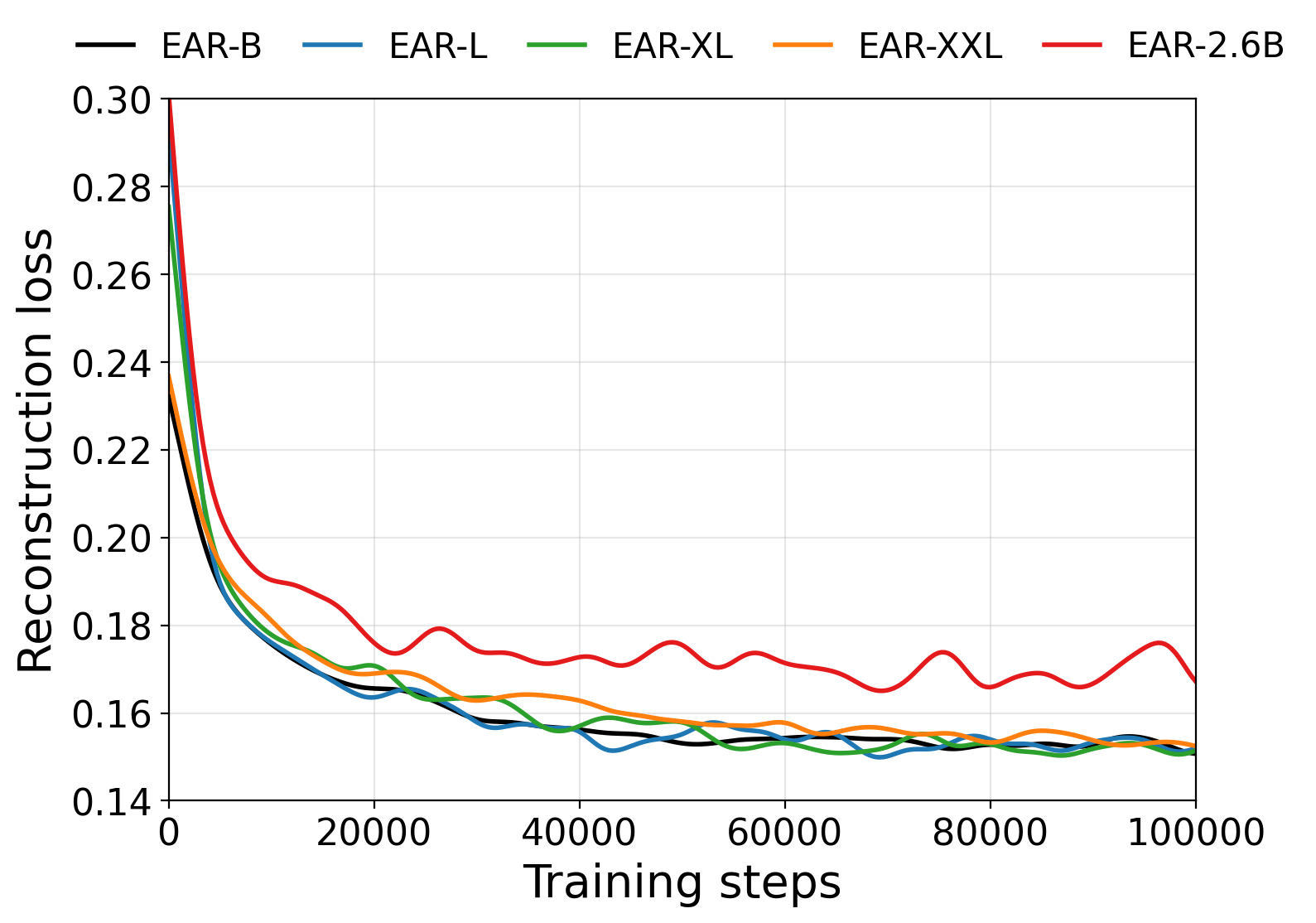}
        \caption{Prior size}
        \label{fig:prior_size}
    \end{subfigure}

    \caption{Larger latent dimension, VAE capacity, and prior capacity for balancing generation and reconstruction.}
    \label{fig:mitigate_ways}
\end{figure}

\textbf{Increasing the latent dimension.}
We first increase the latent dimension to relax the information bottleneck~\cite{vavae}.
As shown in \cref{fig:latent_dim}, larger latent dimensions can tolerate stronger prior weights, but the improvement is limited: even increasing the dimension from 16 to 128 only supports $\lambda_{\mathrm{prior}}\approx0.1$ before reconstruction deteriorates.
Moreover, a larger latent dimension makes the AR prior harder to model.
Thus, increasing latent dimension shifts the trade-off but does not remove the generation--reconstruction conflict.

\textbf{Scaling the VAE.}
We next test whether a stronger VAE can recover details from a more generation-friendly latent space.
Compared with the default VAE ($\mathrm{block}=2$), we increase both encoder and decoder depth ($\mathrm{block}=4$), or increase only the decoder depth.
As shown in \cref{fig:vae_size}, scaling the VAE does not alleviate the conflict and can even hurt reconstruction, suggesting that simply increasing VAE capacity is insufficient.

\textbf{Scaling the generative model.}
Finally, we scale the AR model size to better fit complex latent distributions.
As shown in \cref{fig:prior_size}, a larger prior improves generation but further degrades reconstruction, indicating stronger shaping pressure on the latent space.
Therefore, scaling the prior can amplify the dominance of the generation objective rather than resolving the conflict.

\subsection{Loss Balancing Is Not Enough}
\label{subsec:loss_balancing}

The above results show that increasing the latent dimension, scaling the VAE, or scaling the generative model does not remove the generation--reconstruction conflict. 
We next examine whether better loss balancing can alleviate it. As shown in \cref{fig:abl_1}(a), the prior loss weight $\lambda_{\mathrm{prior}}$ strongly affects this trade-off. 
We therefore test several common weighting schemes with simple tuning, including constant weighting, cosine decay, and PI-style adaptive control~\cite{shao2020controlvae}. 
However, \cref{tab:ablation_pz_method} shows that none consistently achieves a satisfactory balance, highlighting the difficulty of dynamically balancing the two objectives.

To understand this failure, we revisit the asymmetric learning dynamics between reconstruction and generation.
As shown in \cref{fig:rg_visual}, reconstruction is a fast and strongly supervised mapping problem: given latent samples, the decoder receives direct supervision from the target images and can quickly recover visual details. 
In contrast, generation is a slower distribution-modeling problem: the prior must learn the global latent distribution before it can produce realistic samples. 
Therefore, reconstruction can reach good image fidelity within far fewer training steps, whereas generation requires substantially longer training to produce recognizable samples.
Based on this observation, we propose a generation-before-reconstruction training strategy (\textbf{GenFirst}), which treats the conflict as an optimization-order problem rather than a pure loss-balancing problem.
GenFirst lets the generative prior shape a generation-friendly latent space and then strengthens reconstruction to recover visual details within the latent space.

\section{Method}
\label{sec:method}
Building on the analysis in \cref{sec:revisiting_e2e}, we jointly optimize the VAE and the latent generative model under a unified end-to-end objective. 
We first describe the general formulation, then instantiate the latent prior with a continuous autoregressive model~\cite{zheng2025farmer} or SiT~\cite{sit}, and finally present GenFirst, our generation-before-reconstruction training strategy.
\subsection{Stable End-to-End Latent Generative Training}
\label{subsec:e2e_training}
Given an image $x$, the encoder predicts the posterior parameters
$\mu_\phi(x)$ and $\sigma_\phi(x)^2$, from which we sample
$z =
    \mu_\phi(x)
    +
    \sigma_\phi(x)\odot\epsilon_z,
    \ 
    \epsilon_z\sim\mathcal{N}(0,I)$.
The decoder reconstructs the image as $\hat{x}=D_\psi(z)$. 
We optimize the unified end-to-end objective in \cref{eq:e2e_objective}, where the form of $\mathcal{L}_{\mathrm{prior}}$ depends on the generative model.

\textbf{Continuous autoregressive prior.}
For the exact-likelihood autoregressive model, we minimize the latent
negative log-likelihood:
\begin{equation}
    \mathcal{L}^{\mathrm{AR}}_{\mathrm{prior}}
    =
    \mathbb{E}_{x,c,\,z\sim q_\phi(z|x)}
    \left[
        -\log p_\theta(z|c)
    \right],
    \qquad
    p_\theta(z|c)
    =
    \prod_{i=1}^{N}
    p_\theta(z_i|z_{<i},c).
    \label{eq:prior_loss_ar}
\end{equation}

\textbf{SiT prior.}
For SiT, we train a velocity field instead of an explicit likelihood.
Given $\epsilon_t\sim\mathcal{N}(0,I)$ and
$t\sim\mathcal{U}(0,1)$, we construct
\begin{equation}
    z_t=(1-t)\epsilon_t+t z,
    \qquad
    v=z-\epsilon_t,
\end{equation}
and optimize
\begin{equation}
    \mathcal{L}^{\mathrm{SiT}}_{\mathrm{prior}}
    =
    \mathbb{E}_{x,\,z\sim q_\phi(z|x),\,\epsilon_t,\,t}
    \left[
        \left\|
        v_\theta(z_t,t)-(z-\epsilon_t)
        \right\|_2^2
    \right].
    \label{eq:prior_loss_sit}
\end{equation}

Since SiT perturbs $z$ with standard Gaussian noise, the latent value scale must remain bounded; otherwise, an excessively large $z$ would make the injected noise relatively too small, allowing the model to exploit a shortcut to reduce the denoising loss. 
We therefore keep the standard VAE KL regularization with a small weight $10^{-6}$ for the SiT prior loss:
\begin{equation}
    \mathcal{L}^{\mathrm{SiT}}_{\mathrm{E2E}}
    =
    \mathcal{L}_{\mathrm{E2E}}
    +
    10^{-6}
    D_{\mathrm{KL}}
    \left(
        q_\phi(z|x)\,\|\,\mathcal{N}(0,I)
    \right).
    \label{eq:sit_e2e}
\end{equation}
This small KL term is used only for latent-scale regularization and is distinct from the explicit posterior-entropy term in \cref{eq:e2e_objective}, which provides the primary anti-collapse regularization.

\subsection{Continuous Autoregressive Prior}
\label{subsec:ar_prior}

We instantiate the exact-likelihood generative prior with the continuous autoregressive model from FARMER~\cite{zheng2025farmer}, and refer to this end-to-end autoregressive instantiation as \textbf{EAR}.
EAR directly models the VAE latent distribution with an exact likelihood, providing a tractable prior loss for end-to-end training.

Given an image $x$, the VAE encoder produces a latent tensor $z\in\mathbb{R}^{h\times w\times c}$.
We flatten the spatial dimensions of $z$ into a sequence of $N=hw$ continuous tokens: $z = (z_1, z_2, \ldots, z_N), \  z_i\in\mathbb{R}^{c}$.
The autoregressive prior factorizes the latent density as: $p_\theta(z|c)=\prod_{i=1}^{N}p_\theta(z_i|z_{<i},c)$, where $c$ denotes the class or text condition.
We implement this prior with a causal Transformer following FARMER~\cite{zheng2025farmer}.
Condition tokens are placed at the beginning of the sequence, and each latent token attends only to previous condition and latent tokens.
At position $i$, the Transformer hidden state is used to predict the conditional density of the next continuous token $z_i$.
Since each $z_i$ is continuous, we cannot use a discrete softmax head.
Instead, we follow \cite{tschannen2023givt,tschannen2024jetformer,zheng2025farmer} and use a Gaussian mixture model (GMM) head.
Given the hidden state $h_i$, the head predicts mixture weights, means, and variances for a $K$-component diagonal Gaussian mixture:
\begin{equation}
    p_\theta(z_i|z_{<i},c)
    =
    \sum_{k=1}^{K}
    \pi_{i,k}
    \mathcal{N}
    \left(
        z_i;
        m_{i,k},
        \mathrm{diag}(s_{i,k}^2)
    \right).
    \label{eq:gmm_head}
\end{equation}
The prior loss is the autoregressive negative log-likelihood in \cref{eq:prior_loss_ar}.
During end-to-end training, this loss updates both the AR prior and the VAE encoder.
Therefore, EAR not only learns to fit the current latent distribution, but also provides gradients that reshape the VAE latent space into one that is easier for the autoregressive prior to model.

\subsection{Generation-before-Reconstruction Training}
\label{subsec:gbr_training}

Motivated by the asymmetric learning dynamics discussed in \cref{subsec:loss_balancing}, we propose generation-before-reconstruction (\textbf{GenFirst}), a two-stage schedule based on the same end-to-end objective in \cref{eq:e2e_objective}.
Throughout training, we keep the reconstruction, LPIPS, and GAN weights unchanged from standard VAE training, and tie the entropy weight to the prior weight, i.e., $\lambda_{\mathrm{ent}}=\lambda_{\mathrm{prior}}$.
GenFirst uses a larger prior weight in the generation-first stage and a smaller one in the reconstruction-refinement stage:
\begin{equation}
    \lambda_{\mathrm{prior}}^{(1)}
    >
    \lambda_{\mathrm{prior}}^{(2)},
    \qquad
    \lambda_{\mathrm{ent}}^{(s)}
    =
    \lambda_{\mathrm{prior}}^{(s)},
    \quad s\in\{1,2\}.
\end{equation}

In the generation-first stage, the stronger generative objective plays the main role in shaping a generation-friendly latent space, while reconstruction preserves image information and posterior entropy prevents latent collapse.
In the reconstruction-refinement stage, we reduce the prior and entropy weights so that reconstruction receives greater relative influence and recovers visual details without substantially disrupting the generation-friendly latent space formed in the first stage.
In practice, we use a longer generation-first stage for latent-space formation and a shorter reconstruction-refinement stage for recovering visual details.

Unlike cosine schedules or PI-style adaptive weighting~\cite{shao2020controlvae}, GenFirst does not merely adjust loss magnitudes; it explicitly separates latent-space formation from reconstruction refinement according to the asymmetric learning dynamics of generation and reconstruction.

\section{Experiments}
\label{sec:experiments}

\subsection{Experimental Setup}
\label{subsec:exp_setup}
Architecture configurations, dataset details, and stage-specific optimization details are provided in \cref{app:training_details}.

\textbf{Datasets and metrics.}
We evaluate class-conditional generation on ImageNet~\cite{imagenet} at $256\times256$ and $512\times512$ resolutions. 
For text-to-image generation, we train on publicly available datasets~\cite{JourneyDB,midjounery23m,megalith,imagenet,cc12m,commoncatalog,open4o}, and evaluate on GenEval~\cite{geneval} and DPG-Bench~\cite{dpg}.
For ImageNet generation, we report gFID~\cite{fid}, Inception Score (IS)~\cite{is}, Precision, and Recall~\cite{precision} using 50K generated samples. 
For reconstruction, we report rFID, PSNR~\cite{psnr}, LPIPS~\cite{lpips}, and SSIM~\cite{ssim}.

\textbf{EAR.}
EAR combines the VAE architecture from MAR~\cite{mar}, using the $f16d16$ latent setting, with the continuous autoregressive prior from FARMER~\cite{zheng2025farmer}. 
The VAE and AR prior are jointly trained from scratch using our end-to-end objective with GenFirst.
We set $\lambda_{\mathrm{prior}}=\lambda_{\mathrm{ent}}=1$ for the 500-epoch generation-first stage and reduce both weights to $0.25$ for the 140-epoch reconstruction-refinement stage.
We use a batch size of 512 and linearly warm up the learning rate to $1\times10^{-4}$ during the first 10 epochs, then keep it constant.

\textbf{EiT.}
EiT denotes our diffusion-based instantiation of the end-to-end framework.
We use SiT~\cite{sit} for class-conditional ImageNet generation and MMDiT~\cite{mmdit} for text-to-image generation.
Following REPA-E~\cite{repa_e}, training consists of an end-to-end phase followed by a prior-only phase. 
During the end-to-end phase, the VAE and SiT are jointly optimized using our end-to-end objective with GenFirst.
We set $\lambda_{\mathrm{prior}}=\lambda_{\mathrm{ent}}=0.1$ for the 80-epoch generation-first stage and reduce both weights to $0.01$ for the 40-epoch reconstruction-refinement stage.
During the prior-only phase, we freeze the VAE and train SiT for an additional 800 epochs.
Other optimization settings follow REPA-E~\cite{repa_e}.

\textbf{High-resolution generation.}
For both EAR and EiT, we reuse the VAE trained at $256\times256$ resolution and freeze it. We then train only the corresponding AR or SiT generative prior at $512\times512$ resolution.

\textbf{Text-to-image generation.}
We use the same two-phase training protocol for text-to-image generation.
During the end-to-end phase, we consider two variants.
\textit{Version A} jointly trains SiT-XL/2 and the FLUX.1-dev VAE on ImageNet using the same end-to-end protocol as class-conditional EiT.
\textit{Version B} jointly trains a MMDiT-L/2~\cite{mmdit} and the FLUX.1-dev VAE directly on the text-to-image training data.
During the prior-only phase, we freeze the resulting VAE and train the target MMDiT-XL/2 on the text-to-image dataset.
Following UniDDT~\cite{wang2026uniddt} and PixelGen~\cite{ma2026pixelgen}, we train MMDiT at $256\times256$ for 200K steps and then at $512\times512$ for 80K steps. 
We further fine-tune it on OpenAI-4o datasets~\cite{open4o} for 40K steps as $512\times512$.

\subsection{Analysis}
\label{subsec:analysis}
For ease of analysis, unless otherwise specified, we use the exact-likelihood EAR as the default setting, and report gFID and IS without guidance.

\textbf{Does end-to-end training learn a generation-friendly latent space?}
We compare four protocols in \cref{tab:ear_sit_training_settings}:
\textit{Frozen VAE} is the standard two-stage baseline; 
\textit{Naive E2E} propagates the generative loss to the VAE without explicit entropy preservation; \textit{Stable E2E} uses our entropy-preserving objective and GenFirst strategy; 
and \textit{Frozen E2E-trined VAE} freezes the VAE learned by our Stable E2E objective with GenFirst and trains a new prior from scratch.
Naive E2E collapses for both EAR and SiT, whereas Stable E2E remains stable.
More importantly, replacing the reconstruction-trained VAE with the frozen E2E-trained VAE substantially improves a newly trained prior: gFID decreases from $36.33$ to $5.67$ for EAR and from $7.90$ to $3.57$ for SiT.
These results show that stable end-to-end training learns a more generation-friendly latent space.
EAR and SiT exhibit different behaviors during joint training.
For EAR, the tractable likelihood directly models the VAE latent density, so Stable E2E slightly outperforms retraining the prior on the frozen E2E-trained VAE.
In contrast, SiT lacks an explicit likelihood over VAE latents, making joint VAE--SiT optimization less effective than freezing the E2E-trained VAE and training a new SiT prior.

\begin{table}[!h]
\scriptsize
\centering
\caption{
\textbf{Stable end-to-end training learns a generation-friendly latent space.}
EAR is trained until gFID converges, whereas SiT is evaluated after 80 epochs.
Frozen E2E-trained VAE freezes the VAE obtained from Stable E2E and trains a new prior from scratch.
\emph{Collapsed} indicates numerical instability with a NaN training loss.
}

\begin{minipage}{0.48\linewidth}
\centering
\textbf{EAR}\\[2mm]
\setlength{\tabcolsep}{0.8mm}
\begin{tabular}{lccccc}
\toprule
Setting & \makecell{gFID$\downarrow$} & IS$\uparrow$ & rFID$\downarrow$ & PSNR$\uparrow$ & SSIM$\uparrow$ \\
\midrule
Frozen VAE
& 36.33 & 46.55 & \textbf{0.53} & \textbf{26.18} & \textbf{0.72} \\
Naive E2E
& 294.90 & 1.48 & 0.70 & 25.11 & 0.68 \\
Stable E2E (ours)
& \textbf{5.10} & \textbf{145.22} & 1.26 & 21.93 & 0.58 \\
Frozen E2E-trained VAE
& 5.67 & 137.39 & 1.26 & 21.93 & 0.58 \\
\bottomrule
\end{tabular}
\end{minipage}
\hfill
\begin{minipage}{0.48\linewidth}
\centering
\textbf{SiT (80 epochs)}\\[2mm]
\setlength{\tabcolsep}{0.8mm}
\begin{tabular}{lccccc}
\toprule
Setting & \makecell{gFID$\downarrow$} & IS$\uparrow$ & rFID$\downarrow$ & PSNR$\uparrow$ & SSIM$\uparrow$ \\
\midrule
Frozen VAE
& 7.90 & 122.60 & 0.74 & \textbf{25.67} & \textbf{0.72} \\
Naive E2E
& \multicolumn{5}{c}{Collapsed} \\
Stable E2E (ours)
& 92.38 & 13.94 & \textbf{0.57} & 25.13 & 0.71 \\
Frozen E2E-trained VAE
& \textbf{3.57} & \textbf{166.49} & \textbf{0.57} & 25.13 & 0.71 \\
\bottomrule
\end{tabular}
\end{minipage}

\label{tab:ear_sit_training_settings}
\end{table}

\begin{figure}
    \centering
    \includegraphics[width=\linewidth]{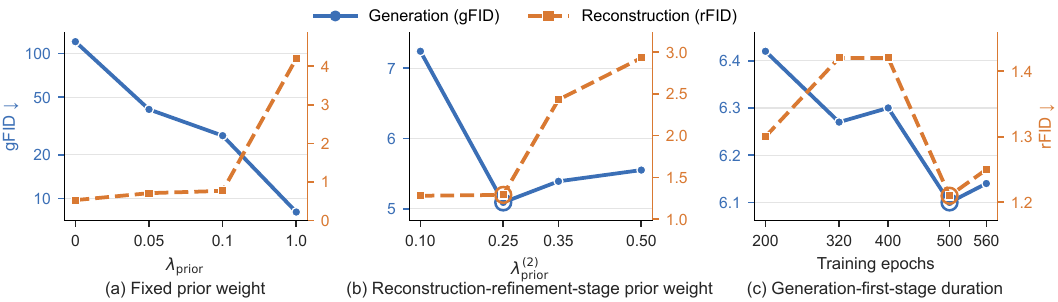}
    \caption{
    \textbf{Generation--reconstruction trade-off and GenFirst schedule ablations.}
We report gFID (blue solid line) and rFID (orange dashed line), where lower is better.
(a) \textbf{Effect of prior loss weight.}
All models are trained with the same architecture and training budget.
No fixed prior weight achieves a favorable generation--reconstruction trade-off.
(b) \textbf{Effect of the reconstruction-refinement-stage prior weight.}
A moderate prior weight in the reconstruction refinement stage gives the best generation--reconstruction trade-off.
(c) \textbf{Effect of the generation-first-stage duration.} gFID decreases with more generation-first-stage training epochs and saturates with further training.
Circled points indicate the default settings.
    }
    \label{fig:abl_1}
\end{figure}

\begin{table}[!h]
\scriptsize
\centering
\caption{
\textbf{Comparison of loss balancing strategies.}
GenFirst denotes our generation-before-reconstruction schedule. This experiment evaluates whether the generation-reconstruction conflict can be solved by simple weighting schedules, or whether the order of optimization matters. 
\emph{Collapsed} indicates numerical instability with a NaN training loss.
}
\setlength{\tabcolsep}{2.1mm}{
\begin{tabular}{cccccc}
\toprule
Strategy
&
gFID (w. CFG)$\downarrow$
&
IS (w. CFG)$\uparrow$
&
rFID$\downarrow$
&
PSNR$\uparrow$
&
SSIM$\uparrow$
\\
\midrule
\textbf{GenFirst}
&  \textbf{2.10} & 246.70	& 1.26 & 21.93 & 0.58 
\\
Constant
&  3.04	& \textbf{251.2}	& 3.27 & 19.52 & 0.48 
\\
Cosine decay
&  5.45	& 136.11	& 0.66 & 23.87 & 0.67
\\
PI adaptive
&  \multicolumn{5}{c}{Collapsed}
\\
Frozen VAE
& 6.97 & 147.04	& \textbf{0.53}	&\textbf{26.18}	&\textbf{0.72}
\\
\bottomrule
\end{tabular}
}
\label{tab:ablation_pz_method}
\end{table}

\textbf{How does prior weighting affect the generation--reconstruction trade-off?} 
As shown in \cref{fig:abl_1}(a), increasing the prior loss weight improves generation but degrades reconstruction. 
When $\lambda_{\mathrm{prior}}$ increases from 0 to 1.0, gFID improves from 120.60 to 8.06 and IS increases from 13.07 to 113.51, showing that stronger prior fitting makes the latent distribution easier to model. 
However, reconstruction quality drops substantially: rFID worsens from 0.53 to 4.20, PSNR decreases from 26.18 to 19.66, and SSIM decreases from 0.72 to 0.45. 
This confirms that the key challenge in end-to-end training is balancing a generation-friendly latent distribution with sufficient information for faithful decoding.

\textbf{Can simple loss balancing strategies resolve the conflict?}
We compare GenFirst with three representative weighting baselines: constant weighting, cosine decay, and PI-based adaptive control~\cite{shao2020controlvae}; full settings are provided in \cref{app:loss_balancing}. 
As shown in \cref{tab:ablation_pz_method}, the tested baselines reach different points on the generation--reconstruction trade-off. 
Constant weighting achieves reasonable generation but degrades reconstruction (gFID $3.04$, rFID $3.27$), whereas cosine decay improves reconstruction at the cost of generation (gFID $5.45$, rFID $0.66$).
The PI-controlled run becomes numerically unstable in our setting. 
GenFirst achieves the best gFID among the evaluated strategies, with a gFID of $2.10$, an IS of $246.70$, and an rFID of $1.26$.
These results suggest that it is difficult to solve the conflict by simply tuning or scheduling the prior weight; the optimization order is critical.

\textbf{Effect of the reconstruction-refinement-stage prior weight.}
We further study how much prior pressure should be kept during the reconstruction-refinement stage.
As shown in \cref{fig:abl_1}(b), reducing the weight from $0.5$ to $0.25$ improves both objectives: gFID decreases from $5.55$ to $5.09$, while rFID decreases from $2.93$ to $1.29$.
When the weight is further reduced to $0.1$, reconstruction improves slightly, but generation degrades to a gFID of $7.24$.
We therefore set $\lambda_{\mathrm{prior}}^{(2)}=0.25$, which provides the best trade-off among the tested values.

\textbf{Effect of the generation-first-stage duration.}
We also vary the first-stage training duration from 200 to 560 epochs.
As shown in \cref{fig:abl_1}(c), the results remain relatively stable across this range, with gFID varying from $6.10$ to $6.42$ and rFID from $1.21$ to $1.42$.
Training for 500 epochs achieves the lowest gFID and rFID among the evaluated settings, while extending training to 560 epochs brings no further improvement.
We therefore use 500 epochs as the default generation-first-stage training epoch.

\begin{wraptable}{r}{0.35\columnwidth}
\vspace{-1.2\baselineskip}
\centering
\caption{
\textbf{Scaling behavior of EAR.}
The VAE is first jointly trained with an EAR-L prior under GenFirst; it is then frozen while we separately vary the EAR model size and the number of GMM components.
}
\label{tab:ear_scaling}
\scriptsize
\setlength{\tabcolsep}{2.2pt}
\textbf{(a) EAR model scale.}\\[2pt]
\begin{tabular*}{0.95\linewidth}{
@{\extracolsep{\fill}}
cccc
@{}
}
\toprule
Model & \#Params & gFID$\downarrow$ & IS$\uparrow$ \\
\midrule
B    & 174M & 13.85 & 79.00  \\
L    & 477M & 5.67  & 137.39 \\
XL   & 775M & 3.91  & 163.71 \\
XXL  & 1.4B & 3.19  & 181.99 \\
\bottomrule
\end{tabular*}

\vspace{5pt}
\textbf{(b) GMM components.}\\[2pt]
\begin{tabular*}{0.95\linewidth}{
@{\extracolsep{\fill}}
ccc
@{}
}
\toprule
GMM $K$ & gFID$\downarrow$ & IS$\uparrow$ \\
\midrule
32   & 6.08 & 132.70 \\
64   & 5.67 & 137.39 \\
128  & 5.64 & 137.24 \\
1024 & 5.01 & 144.63 \\
4096 & 5.00 & 144.99 \\
\bottomrule
\end{tabular*}
\vspace{-1.5\baselineskip}
\end{wraptable}

\textbf{Scalability of the EAR prior.}
We first jointly train the VAE with an EAR-L prior using GenFirst.
We then freeze the VAE and train new EAR priors of different sizes from scratch on its latents, using a lightweight 6-layer normalizing-flow adapter~\cite{tarflow,tschannen2023givt}.
As shown in \cref{tab:ear_scaling}(a), scaling the prior from 174M to 1.4B parameters consistently reduces gFID from $13.85$ to $3.19$, while IS increases from $79.00$ to $181.99$.
This shows that GenFirst-trained latents continue to benefit from larger generative priors.

\textbf{Effect of GMM capacity.}
Using the same frozen VAE, we fix the autoregressive backbone to EAR-L and train new priors from scratch with different numbers of GMM components $K$.
As shown in \cref{tab:ear_scaling}(b), increasing $K$ from 32 to 1024 improves gFID from $6.08$ to $5.01$ and IS from $132.70$ to $144.63$.
Further increasing $K$ to 4096 brings only marginal gains, indicating that the benefit largely saturates beyond $K=1024$ in this setting.

\begin{table*}[!t]
\centering

\begin{minipage}[t]{0.6\textwidth}
\vspace{0pt}
\centering
\captionof{table}{
\textbf{Comparison with REPA-E.}
We compare EiT with REPA and REPA-E under different VAE initialization
settings. All models are evaluated after 80 training epochs.
}
\label{tab:vae_initialization}

\scriptsize
\setlength{\tabcolsep}{3.0pt}
\begin{tabular}{lcccc}
\toprule
\textbf{Method}
& \textbf{gFID}$\downarrow$
& \textbf{IS}$\uparrow$
& \textbf{Prec.}$\uparrow$
& \textbf{Rec.}$\uparrow$ \\
\midrule
REPA~\cite{repa}
& 7.90 & 122.60 & 0.70 & \textbf{0.65} \\

REPA-E (scratch)~\cite{repa_e}
& 4.34 & 154.30 & 0.75 & 0.63 \\

REPA-E (SD-VAE init.)~\cite{repa_e}
& 4.07 & 161.80 & 0.76 & 0.62 \\

REPA-E (VA-VAE init.)~\cite{repa_e}
& 3.46 & 159.80 & 0.77 & 0.63 \\

\textbf{EiT} (scratch)
& 3.78 & 161.42 & \textbf{0.78} & 0.61 \\

\textbf{EiT} (SD-VAE init.)
& 3.57 & 166.49 & 0.77 & 0.62 \\

\textbf{EiT} (VA-VAE init.)  
& \textbf{2.79} & \textbf{175.40} & \textbf{0.78} & 0.63 \\
\bottomrule
\end{tabular}
\end{minipage}
\hfill
\begin{minipage}[t]{0.37\textwidth}
\vspace{0pt}
\centering
\captionof{table}{
\textbf{Effect of an optional normalizing flow.}
The VAE is frozen when training each AR prior.
}
\label{tab:optional_nf}

\scriptsize
\setlength{\tabcolsep}{3.pt}
\begin{tabular}{lcc}
\toprule
\textbf{Setting}
& \textbf{gFID}$\downarrow$
& \textbf{IS}$\uparrow$ \\
\midrule
Frozen VAE + AR
& 110.10 & 14.79 \\

Frozen VAE + NF + AR
& 36.33 & 46.55 \\
\midrule
Stable E2E VAE + AR
& 5.27 & 149.28 \\

Stable E2E VAE + NF + AR
& \textbf{5.12} & \textbf{151.17} \\
\bottomrule
\end{tabular}
\end{minipage}

\end{table*}

\textbf{Comparison with REPA-E.}
We compare EiT, which jointly optimizes SiT~\cite{sit} and the VAE using our end-to-end objective with GenFirst, with REPA~\cite{repa} and REPA-E~\cite{repa_e} under the same 80-epoch training budget in \cref{tab:vae_initialization}.
Compared with REPA-E, EiT consistently achieves lower gFID across VAE initializations: $3.78$ vs.\ $4.34$ from scratch, $3.57$ vs.\ $4.07$ with SD-VAE initialization, and $2.79$ vs.\ $3.46$ with VA-VAE initialization. 
These results show that our end-to-end formulation with GenFirst further improves generation beyond REPA-E.

\textbf{Optional normalizing flow.}
Following FARMER~\cite{zheng2025farmer}, we examine whether an additional
normalizing flow (NF) is still needed for continuous AR modeling.
As shown in \cref{tab:optional_nf}, adding NF to the frozen-VAE setting improves
gFID from $110.10$ to $36.33$, indicating that NF transforms the
reconstruction-optimized latent distribution into one that is easier for the AR
prior to model.
After end-to-end VAE training, however, the gain becomes marginal.
This suggests that end-to-end training already produces a latent space well
suited to continuous AR modeling, making the additional flow largely optional.

\begin{figure}
    \centering
    \includegraphics[width=\linewidth]{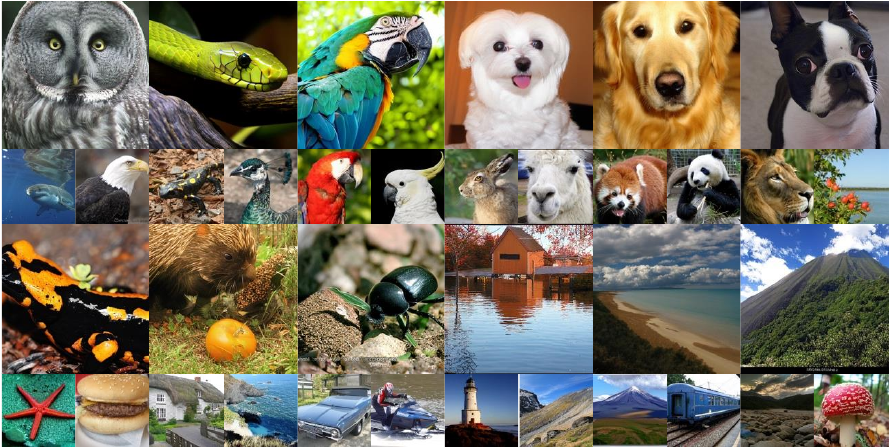}
    \caption{Class-to-image samples from EiT (row 1,2) and EAR (row 3,4) with guidance(CFG scale=4).}
    \label{fig:vis}
\end{figure}

\begin{figure}
    \centering
    \includegraphics[width=\linewidth]{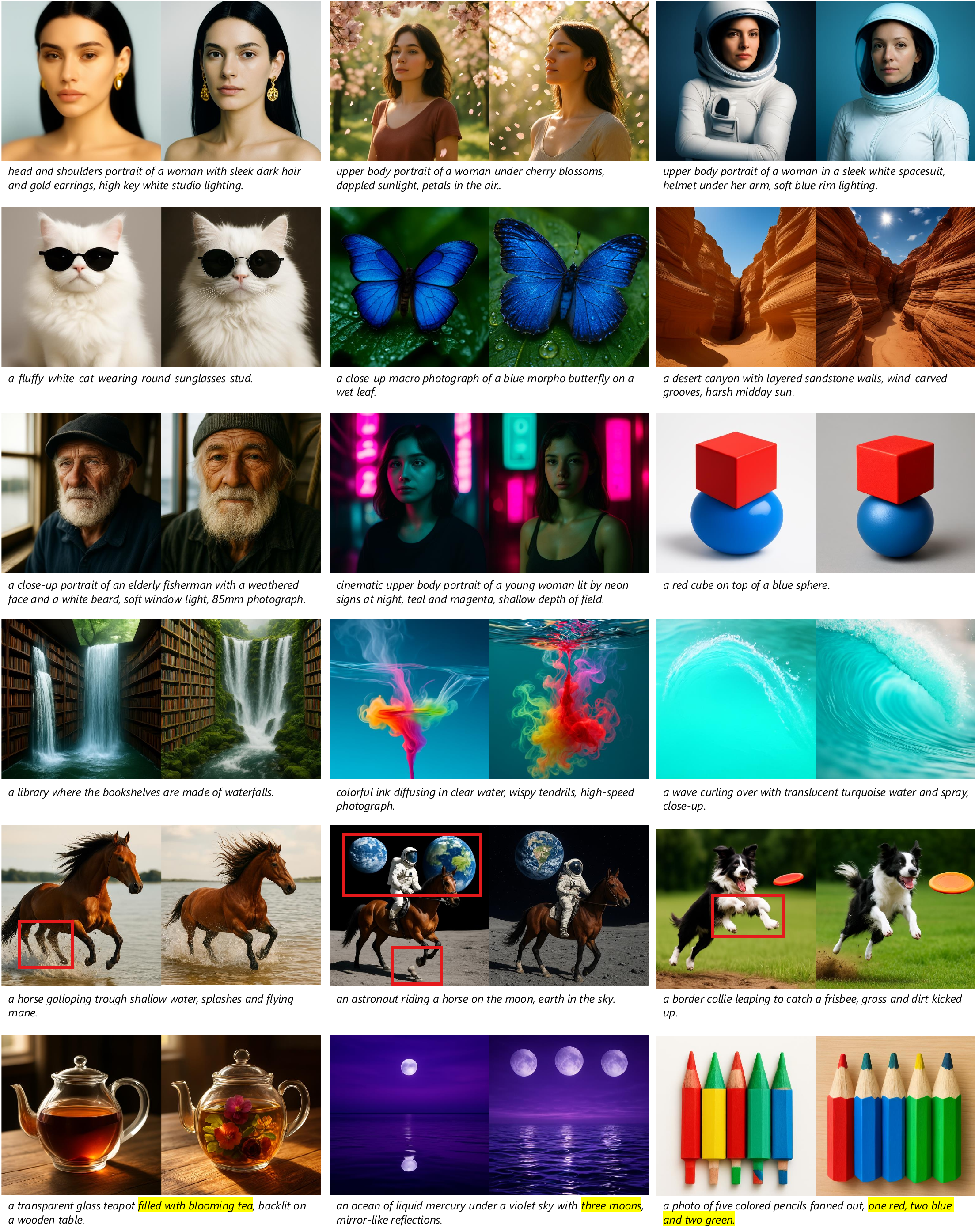}
    \caption{\textbf{Qualitative comparison of text-to-image generation.} Both settings use the same model and training data; the \textbf{left} uses the FLUX.1-dev VAE and the \textbf{right} our end-to-end trained VAE. End-to-end training improves perceptual quality and fine-grained textures (rows 1--3), produces more plausible and coherent images (rows 4--5), and improves text alignment (row 6).}
    \label{fig:vis2}
\end{figure}

\begin{table}[]
\scriptsize 
    \centering
    \caption{\textbf{Class-conditional performance on ImageNet 256$\times$256.} $*$ indicates the results obtained by sampling images with low-temperature strategy.}
    \setlength{\tabcolsep}{1.5mm}{
\begin{tabular}{l cc cccc cccc}
\toprule
\multirow{2}{*}{Method} & \multirow{2}{*}{\makecell{Epochs}} & \multirow{2}{*}{\#Params} & \multicolumn{4}{c}{W/o guidance} & \multicolumn{4}{c}{W/ guidance} \\
\cmidrule(lr){4-7} \cmidrule(lr){8-11}
 & & & gFID$\downarrow$ & IS$\uparrow$ & Prec.$\uparrow$ & Rec.$\uparrow$ & gFID$\downarrow$ & IS$\uparrow$ & Prec.$\uparrow$ & Rec.$\uparrow$ \\
\midrule
\multicolumn{11}{l}{\textit{Pixel Space}} \\
\arrayrulecolor{black!30}\midrule
ADM~\citep{adm} &  400  &  554M  & 10.94 &  101.0 & 0.69 & 0.63 & 3.94 & 215.8 & \textbf{0.83} & 0.53\\
RIN~\citep{jabri2022scalable} &  480  &  410M & 3.42  & 182.0  &  -   &  -   &  -   &   -    &  -   &  -  \\
PixelFlow~\citep{chen2025pixelflow} & 320 & 677M &  - & -   &   -  &  -   & 1.98 & 282.1 & 0.81 & 0.60 \\
PixNerd~\citep{wang2025pixnerd} & 160 & 700M &  -   &  -       &  - &  -  &  2.15 & 297.0 & 0.79 & 0.59 \\
SiD2~\citep{hoogeboom2024simpler} &  1280   &   -   & -   &  - &  -   &  -   &  1.38   &   -    &  -   &  - \\
TARFlow~\citep{tarflow} & 320 & 1.4B & -  & - & - & - & 4.69 & - & - & - \\
JetFormer~\citep{tschannen2024jetformer} & 500 & 2.8B  & - & - & - & -  & 6.64 & - & 0.69 & 0.56 \\
FARMER~\citep{zheng2025farmer} & 320 & 1.9B &  - & - & - & - & 3.60 & 269.2 & 0.81 & 0.51\\
JiT~\citep{jit} & 600 & 2B &  - & - & - & - & 1.82 & 292.6 & 0.79 & 0.62\\
PixelDiT~\citep{pixeldit} & 320 & 797M & - & - & - & - & 1.61 & 292.7 & 0.78 & 0.64\\
DeCo~\citep{deco} & 600 & 682M & - & - & - & - & 1.69 & 304.0 & 0.79 & 0.63 \\
\arrayrulecolor{black}\midrule
\multicolumn{11}{l}{\textit{Latent Autoregressive}} \\
\arrayrulecolor{black!30}\midrule
GIVT~\citep{tschannen2023givt} & 500 & 1.67B & - & - & - & - & 2.59 & - &  0.81 & 0.57\\
VAR~\citep{var} &  350  &  2.0B  & 1.92 & \textbf{323.1}  & \textbf{0.82} & 0.59 & 1.73 & \textbf{350.2} & 0.82 & 0.60\\
MAR~\citep{mar} &  800  &  943M   & 2.35 &  227.8 & 0.79 & 0.62 & 1.55 & 303.7 & 0.81 & 0.62\\
xAR~\citep{xar} &  800  & 1.1B  &  - & - & - & - & 1.24 & 301.6 & \textbf{0.83} & 0.64\\
\arrayrulecolor{black}\midrule
\multicolumn{11}{l}{\textit{Latent Diffusion}} \\
\arrayrulecolor{black!30}\midrule
DiT~\citep{dit} & 1400 & 675M &  9.62 & 121.5 & 0.67 & 0.67 & 2.27 & 278.2 & \textbf{0.83} & 0.57 \\
MaskDiT~\citep{maskdit} & 1600 & 675M &  5.69 & 177.9 & 0.74 & 0.60 & 2.28 & 276.6 & 0.80 & 0.61 \\
SiT~\citep{sit} & 1400 & 675M &  8.61 & 131.7 & 0.68 & 0.67 & 2.06 & 270.3 & 0.82 & 0.59 \\
MDTv2~\citep{gao2023mdtv2} & 1080 & 675M &  - & - & - & - & 1.58 & 314.7 & 0.79 & 0.65 \\
REPA~\citep{repa} & 800 & 675M &  5.78 & 158.3 & 0.70 & 0.68 & 1.29 & 306.3 & 0.79 & 0.64 \\
VA-VAE~\citep{vavae} & 800 & 675M &  2.17 & 205.6 & 0.77 & 0.65 & 1.35 & 295.3 & 0.79 & 0.65 \\
DDT~\citep{decoupled_dit} & 400 & 675M &  6.27 & 154.7 & 0.68 & \textbf{0.69} & 1.26 & 310.6 & 0.79 & 0.65\\
RAE~\citep{rae} & 800 & 839M &  1.51 & 242.9 & 0.79 & 0.63 & 1.13 & 262.6 & 0.78 & \textbf{0.67} \\
 \arrayrulecolor{black}\midrule
\multicolumn{11}{l}{\textit{End-to-End Generative models}} \\
\arrayrulecolor{black!30}\midrule
SimFlow~\citep{simflow} & 160 & 1.4B &  13.72 & 105.2 & 0.67 & 0.62 & 2.15 & 276.8 & 0.83 & 0.57 \\
REPA-E~\citep{repa_e}  & 800 & 675M &  1.69 & 219.3 & 0.77 & 0.67 & 1.12 & 302.9 & 0.79 & 0.66 \\
\rowcolor{blue!8}\textbf{EAR (Ours) (w/o GenFirst)} & 500 & 312M & 5.17 & 147.5 & 0.70 &
0.59 & 3.04 & 251.2 & 0.73 & 0.55\\
\rowcolor{blue!8}\textbf{EAR (Ours)} & 500+140 & 312M & 6.10/3.83* & 138.4/164.0* & 0.65 & 0.64 & 2.10 & 246.7 & 0.72 & 0.59 \\
\rowcolor{blue!8}\textbf{EiT (Ours)} & 80 & 675M & 2.79 & 175.4 & 0.78 & 0.63 & 1.38 & 271.3 & 0.79 & 0.63 \\
\rowcolor{blue!8}\textbf{EiT (Ours)} & 480 & 675M & 1.60 & 216.8	& 0.78	& 0.65 & 0.988 & 270.7 & 0.79	& \textbf{0.67}\\
\rowcolor{blue!8}\textbf{EiT (Ours)} & 800 & 675M & \textbf{1.45}	& 226.7	& 0.79	& 0.65 & \textbf{0.969} & 283.4 & 0.79	& \textbf{0.67}\\
\arrayrulecolor{black}\bottomrule
\end{tabular}
}
\label{tab:imagenet256_sota}
\end{table}

\begin{table}[]
\scriptsize 
    \centering
    \caption{\textbf{Class-conditional performance on ImageNet 512$\times$512.} }
    \setlength{\tabcolsep}{2.0mm}{
\begin{tabular}{lcccccc}
\toprule
Method & Epochs & \#Params & gFID$\downarrow$  &  IS$\uparrow$  & Prec.$\uparrow$ & Rec.$\uparrow$ \\
\midrule
BigGAN-deep~\citep{biggan} & - & 158M & 8.43 & 177.9 & 0.88 & 0.29 \\
StyleGAN-XL~\citep{sauer2022stylegan} & - & -  & 2.41 &  267.8 & 0.77 & 0.52 \\
\arrayrulecolor{black}\midrule
VAR~\citep{var} & 350 & 2.3B & 2.63 & 303.2 & - & - \\
MAGVIT-v2~\citep{magvitv2} & 1080 & 307M & 1.91 & 324.3 & - & - \\
MAR~\citep{mar} & 800 & 481M & 1.73 & 279.9 & - & - \\
XAR~\citep{xar} & 800 & 608M & 1.70 & 281.5 & - & - \\
\arrayrulecolor{black}\midrule
ADM~\citep{adm} & 871+210 & 731M & 3.85 &  221.7 & 0.84 & 0.53\\
SiD2~\citep{hoogeboom2024simpler}  & - & - & 1.50   &    -     &   -  &   -  \\
PixNerd~\cite{wang2025pixnerd} & 340 & 700M & 2.84 & 245.6 & 0.80 & 0.59 \\
JiT~\cite{jit} & 600 & 2B & 1.78 & 306.8 & - & - \\
PixelDiT~\cite{pixeldit} & 850 & 797M & 1.81 & 278.6 & 0.78 & 0.67 \\
DeCo~\cite{deco} &  340 & 682M & 2.22 & 290.0 & 0.80 & 0.60 \\
\arrayrulecolor{black!30}\midrule
DiT~\citep{dit} & 600 & 674M & {3.04} &  {240.8} & 0.84 & 0.54 \\
SiT~\citep{sit} & 600 & 674M & 2.62 &   252.2 & 0.84 & 0.57 \\
REPA~\citep{repa} & 200 & 675M & 2.08 &  274.6 & 0.83 & 0.58 \\
Unified Latents~\cite{heek2026unified} & - & - & 1.31 & - & - & - \\
DDT~\citep{decoupled_dit} & 356 & 675M & 1.28 &  305.1 & 0.80 & 0.63 \\
EDM2~\citep{edm2} & - & 1.5B &1.25 & -& -& -\\
RAE~\citep{rae} & 400 & 839M & 1.13 & 259.6 & 0.80 & 0.63 \\
STARFlow~\citep{starflow} & - & 1.4B & 3.00 & -& -& - \\
\arrayrulecolor{black}\midrule
SimFlow + REPA-E & 160 & 1.4B & 2.74 & 304.9 & 0.81 & 0.57 \\
\rowcolor{blue!8} \textbf{EAR (Ours)} & 320 & 940M & 1.87 & 280.0 & 0.82 & 0.58 \\
\rowcolor{blue!8} \textbf{EiT (Ours)} & 80 & 675M & 1.63 & 278.9 & 0.81 & 0.61\\
\rowcolor{blue!8} \textbf{EiT (Ours)} & 400 & 675M & 1.23 & 286.1 & 0.80 & 0.63 \\
\arrayrulecolor{black}\bottomrule
\end{tabular}
}
\label{tab:imagenet512_sota}
\end{table}

\begin{table*}[t]
\setlength\tabcolsep{4pt}
\centering
\scriptsize 
\caption{Quantitative results on GenEval and DPG-Bench benchmarks. Best results are highlighted in \textbf{bold}. $\dagger$ indicates generation with prompt rewriting. $*$ indicates replacing the original CLIP~\cite{clip} text encoder with Qwen3-1.7B~\cite{bai2025qwen3}.}
\begin{tabular}{lccccccccc}
\toprule
\multirow{2}{*}{Method} & \multirow{2}{*}{\#Params} & \multicolumn{7}{c}{GenEval} & \multirow{2}{*}{DPG-Bench} \\
\cmidrule(lr){3-9}
 & & Single-Obj & Two-Obj & Count & Color & Position & Attr & Overall & \\
\midrule
PixArt~\cite{pixart_sigma}  &   4.3B+0.6B     & 0.98  & 0.50  & 0.44  & 0.80  & 0.08  & 0.07  &  0.48 & 71.11 \\
DALL-E 2\cite{dalle2}  & - & 0.94 & 0.66 & 0.49 & 0.77 & 0.10 & 0.19 & 0.52 & - \\
Show-o~\cite{show_o}   &  1.3B     & 0.95  & 0.52  & 0.49  & 0.82  & 0.11  & 0.28  &  0.53 & - \\
Emu3-Gen~\cite{emu3} &   8B    & 0.98  & 0.71  & 0.34  & 0.81  & 0.17  & 0.21  &  0.54 & 80.60 \\
SDXL\cite{ldm}  & 0.81B+2.6B & 0.98 & 0.74 & 0.39 & 0.85 & 0.15 & 0.23 & 0.55 & 74.65  \\
Janus\cite{janus} & 1.3B & 0.97 & 0.68 & 0.30 & 0.84 & 0.46 & 0.42 & 0.61 & 79.68\\
JanusFlow~\cite{ma2025janusflow} &  1.3B    & 0.97  & 0.59  & 0.45  & 0.83  & 0.53  & 0.42  &  0.63 &  80.09 \\
FLUX.1 [Dev]~\cite{flux1}    &  4.8B+12B    & 0.98  & 0.81  & 0.74  & 0.79  & 0.22  & 0.45  &  0.66 & - \\
DALL-E 3\cite{dalle3}  & - & 0.96 & 0.87 & 0.47 & 0.83 & 0.43 & 0.45 & 0.67 & 83.50 \\
SD3-Medium\cite{mmdit} & 5.5B+2B & 0.99 & 0.94 & 0.72 & 0.89 & 0.33 & 0.60 & 0.74 & 84.08 \\
Show-o2-7B~\cite{show_o2} & 7B & 1.00 & 0.87 & 0.58 & 0.92 & 0.52 & 0.62 & 0.76$\dagger$ & 86.14 \\
MetaQuery-XL\cite{metaquery} & 7B & - & - & - & - & - & - & 0.80$\dagger$ & 82.05 \\
Janus-Pro-7B~\cite{chen2025januspro7b} &  7B  & 0.99  & 0.89  & 0.59  & 0.90  & 0.79  & 0.66  &  0.80  & 84.19\\
Z-Image-Turbo~\cite{team2025zimage} & 4B+6B & 1.00 & 0.95 & 0.77 & 0.89 & 0.65 & 0.68 & 0.82 & 84.86 \\
BLIP3-o-8B\cite{chen2025blip3} & 8B & - & - & - & - & - & - & 0.84 & 81.60 \\
FLUX.2 [Dev]~\cite{flux-2-2025}    & 24B+32B & 1.00 & \textbf{0.99} & 0.79 & 0.93 & 0.73 & 0.78 & 0.87 & \textbf{87.57}\\
Qwen-Image~\cite{qwenimage} &  7B+20B   & 0.99  & 0.92  & \textbf{0.89}  & 0.88  & 0.76  & 0.77  &  0.87 & 88.32 \\ 
BAGEL-7B\cite{bagel} & 7B & 0.98 & 0.95 & 0.84 & 0.95 & 0.78 & 0.77 & 0.88$\dagger$ & -\\
\midrule
\rowcolor{blue!8} \textbf{EiT (Ours)}  & 0.1B+1.3B & \textbf{1.00} & 0.94 & 0.83 & 0.94 & 0.87 & 0.82 & \textbf{0.90} & 82.60
    \\
\rowcolor{blue!8} \textbf{EiT (Ours)$^{*}$}  & 1.7B+1.3B & 0.99 & 0.94 & 0.73 & \textbf{0.96} & \textbf{0.89} & \textbf{0.84} & 0.89 & 84.65
    \\
\bottomrule
\end{tabular}
\label{tab:GenEval}
\end{table*}

\subsection{Main Results}
\label{subsec:main_results}

\textbf{Class-conditional ImageNet generation.}
We report results on ImageNet $256\times256$ and $512\times512$ in
\cref{tab:imagenet256_sota,tab:imagenet512_sota}. We make three main
observations.
First, GenFirst benefits autoregressive generation. EAR uses only 312M parameters and achieves a guided gFID of $2.10$, outperforming substantially larger continuous AR baselines, including GIVT ($2.59$), FARMER ($3.60$), and JetFormer ($6.64$). 
Second, using our end-to-end objective with GenFirst, EiT achieves a guided gFID of $\mathbf{0.988}$ on ImageNet $256\times256$ after only 480 training epochs. 
To our knowledge, EiT is the first diffusion model reported to reach an FID below $1.0$ without using the Fréchet Distance (FD) loss~\cite{fd_loss}.
With 800 epochs, EiT further achieves an unguided gFID of $\mathbf{1.45}$, the
best result among the diffusion models compared in
\cref{tab:imagenet256_sota}.
Third, by jointly shaping the VAE latent space with the generative objective, GenFirst makes the latent space easier to model and substantially accelerates generative modeling. 
After only 80 epochs, EiT achieves an unguided gFID of $2.79$, compared with $7.90$ for REPA and $3.46$ for REPA-E. 
These results demonstrate that the learned latent space supports both exact-likelihood autoregressive and flow-matching generative priors.

At ImageNet $512\times512$, EAR achieves a gFID of $1.87$, while EiT reaches $1.23$ after 400 epochs, showing that the end-to-end trained VAE at ImageNet $256\times256$ transfers effectively to higher-resolution generation.

\begin{figure}
    \centering
    \includegraphics[width=\linewidth]{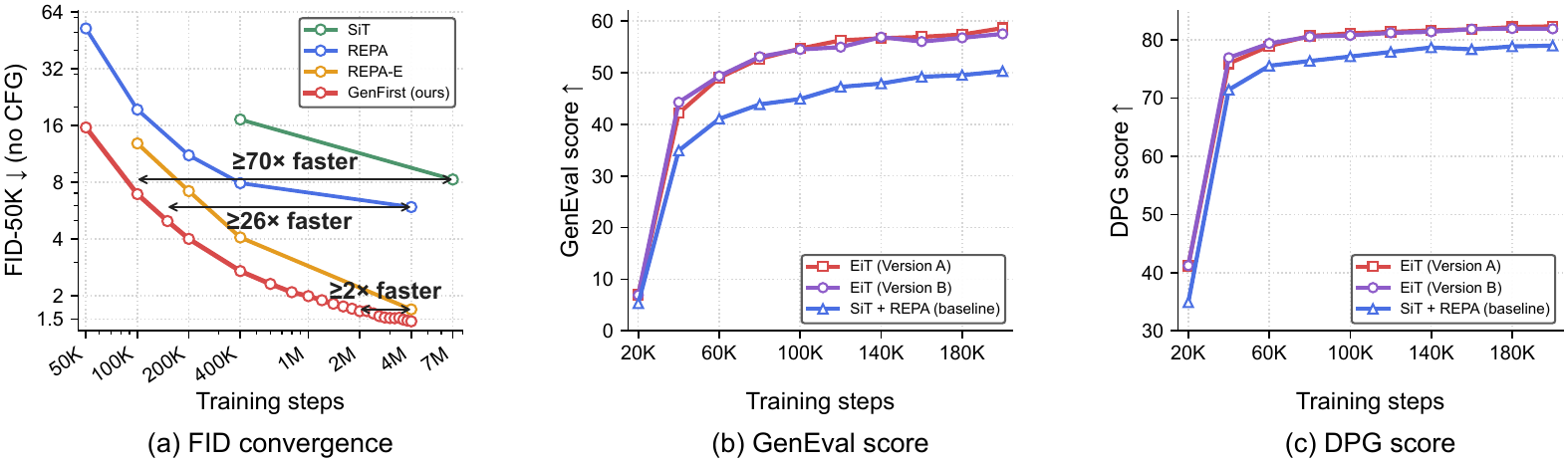}
    \caption{
\textbf{End-to-end training using GenFirst accelerates convergence across generation tasks.}
(a) FID-50K without classifier-free guidance on class-conditional ImageNet generation.
(b--c) GenEval and DPG scores for EiT Versions A/B and the SiT+REPA baseline on text-to-image generation.
Both EiT variants surpass the baseline's 200K-step performance by 80K steps.
Lower FID and higher GenEval and DPG scores indicate better performance.
}
    \label{fig:training_efficiency}
\end{figure}

\textbf{Text-to-image generation.}
For text-to-image generation, the 1.3B EiT achieves an overall GenEval~\cite{geneval} score of $\mathbf{0.90}$, outperforming large-scale text-to-image models such as FLUX.2-dev~\cite{flux-2-2025} and Qwen-Image~\cite{qwenimage}, as well as unified multimodal models, including MetaQuery~\cite{metaquery} and BAGEL~\cite{bagel}.
EiT also achieves a DPG-Bench~\cite{dpg} score of $82.60$, which further improves to 84.65 when replacing the original CLIP text encoder with the stronger Qwen3-1.7B~\cite{bai2025qwen3}.
These results demonstrate strong compositional generation across object composition, counting, color, spatial relations, and attributes.

\textbf{Training efficiency.}
\cref{fig:training_efficiency} shows that GenFirst substantially accelerates convergence across both class-conditional and text-to-image generation.
Specifically, we first jointly train the VAE and SiT using our end-to-end objective with GenFirst; we then freeze the resulting VAE and train a new SiT on this latent space.
For class-conditional generation, GenFirst reaches FID 6.91 at 100K steps and FID 4.97 at 150K steps, outperforming the final results of SiT and REPA with at least $70\times$ and $26\times$ fewer training steps, respectively.
It further matches REPA-E at approximately 2M steps, corresponding to a $2\times$ reduction in training cost, and reaches an FID of 1.45 at 4M steps.
The same trend transfers to text-to-image generation.
Both EiT variants exceed the GenEval and DPG scores of the 200K-step SiT+REPA baseline by 80K steps, while continuing to improve with additional training.
These results show that end-to-end trained latent spaces accelerate the convergence of generative models.

\textbf{Qualitative results.}
\cref{fig:vis} presents class-conditional samples generated by EAR and EiT. 
Both models produce diverse and visually coherent images with fine-grained details and strong correspondence to the input conditions.
\cref{fig:vis2} compares text-to-image generation using the same MMDiT~\cite{mmdit} and training data, differing only in the VAE;
the left uses the FLUX.1-dev VAE and the right our end-to-end trained VAE. 
End-to-end training improves text-to-image generation in three aspects.
First, it substantially improves perceptual quality and fine-grained textures.
For example, the fisherman exhibits more realistic skin details and beard texture, while the cat shows finer and more natural fur; similar improvements are also visible in the butterfly, sandstone, and reflective material examples.
Second, it improves generation plausibility and structural coherence.
For example, the baseline generates a horse with three hind legs, whereas our end-to-end trained VAE produces a more anatomically plausible structure.
Third, our end-to-end trained VAE improves text alignment.
For example, it more faithfully captures the blooming tea inside the transparent teapot and correctly renders the specified count of three moons.

\subsection{Further Extensions}
\label{exp:further_extensions}
\begin{figure}
    \centering
    \includegraphics[width=\linewidth]{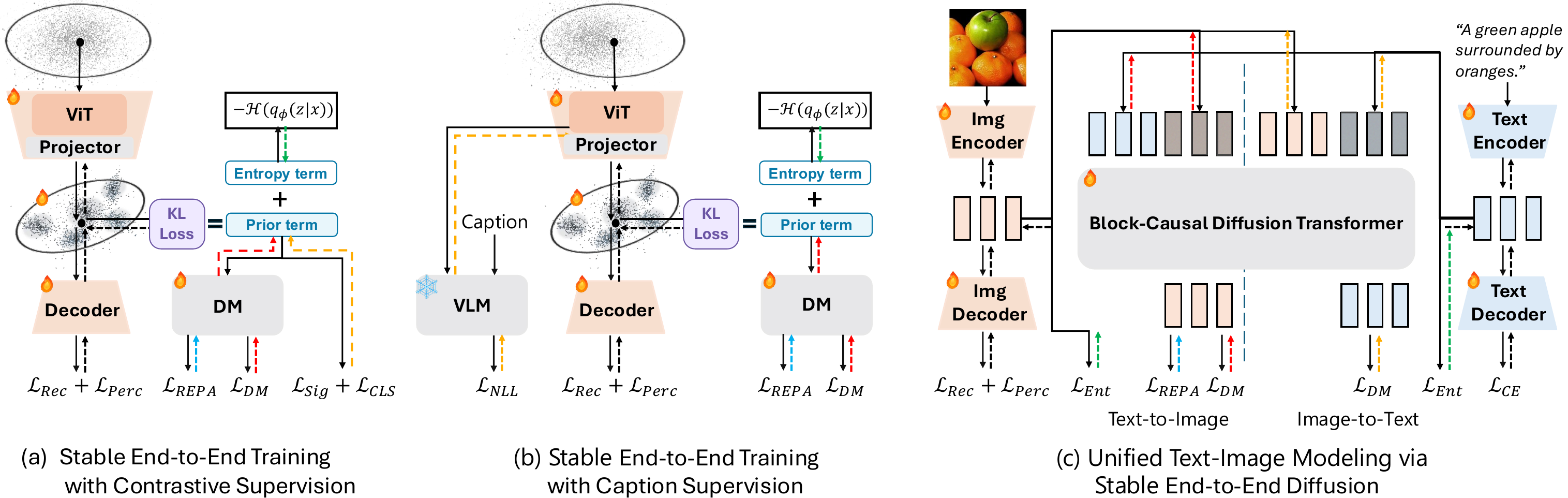}
    \caption{
\textbf{Further extension experiments.}
(a) Stable end-to-end training with latent-level SigLIP supervision for learning a shared latent space for generation and representation learning.
(b) Stable end-to-end training with caption-based VLM-NLL supervision applied to intermediate visual features.
(c) Unified text--image modeling with a block-causal diffusion transformer, where continuous text and image latents are jointly optimized for bidirectional text-to-image and image-to-text generation.
}
    \label{fig:further_exp}
\end{figure}

\begin{table}[t]
\centering
\caption{
Reconstruction and generation performance, and representation quality of different visual representations.
}
\label{tab:und_ablation}
\setlength\tabcolsep{4pt}
\centering
\scriptsize
\begin{tabular}{lccccccc}
\toprule
 \multirow{2}{*}{Encoder} & \multirow{2}{*}{Setting} & \multicolumn{2}{c}{Reconstruction} & \multicolumn{2}{c}{Generation} & \multirow{2}{*}{Top-1 Acc.$\uparrow$} \\
 \cmidrule(lr){3-4} \cmidrule(lr){5-6} 
 & & rFID$\downarrow$ & PSNR$\uparrow$ & GenEval$\uparrow$ & DPG-Bench$\uparrow$ & \\
\midrule
FLUX.1-dev VAE & Pretrained & \textbf{0.18} & \textbf{31.59} & -- & -- & 3.1 \\
Qwen3-VL ViT & Pretrained & -- & -- & -- & -- & 79.64\\
Qwen3-VL ViT+projector & E2E & 0.42 & 27.79 & 0.89 & 83.70 & 77.71\\
Qwen3-VL ViT+projector & E2E + SigLIP supervision &  0.39 & 27.64 & \textbf{0.90} & \textbf{83.76} & 79.25\\
Qwen3-VL ViT+projector & E2E + VLM-NLL supervision & 0.46 & 	
27.19 & 0.89 & 83.49 & \textbf{81.69}\\
\bottomrule
\end{tabular}
\end{table}

\textbf{Toward a shared latent space for generation and representation learning.}
Our experiments above focus on end-to-end training where the latent space is shaped by the generative prior.
Here, we further investigate whether the same framework can learn a shared latent space for both generation and representation learning, such that the latent space is generation-friendly, semantically rich, and supports high-quality reconstruction.
As shown in \cref{fig:further_exp}, we explore two ways of introducing an representation-learning objective into the text-to-image end-to-end training setting described above.
We replace the FLUX.1-dev VAE~\cite{flux1} encoder with the Vision Transformer (ViT)~\cite{vit} from Qwen3-VL~\cite{bai2025qwen3} followed by a dimension-reduction projector that predicts the posterior mean and variance.
We randomly initialize the decoder, freeze the encoder, and first train the decoder for 100K steps for image reconstruction before end-to-end training.

In the first setting (\cref{fig:further_exp}(a)), we directly impose representation supervision on the latent representation.
Specifically, we apply a SigLIP loss~\cite{zhai2023sigmoid,tschannen2025siglip} across samples within each batch together with a classification-based supervision loss~\cite{huang2024classification} on the latent features.
This design allows the representation-learning objective to directly shape the same latent representation used by the generative model.
In the second setting (\cref{fig:further_exp}(b)), we use the intermediate ViT features from the encoder for representation supervision.
We feed these visual features together with the corresponding image captions into Qwen3-VL and minimize the negative log-likelihood (NLL) of the caption tokens~\cite{tschannen2023image}.
Unlike contrastive learning, which typically benefits strongly from large batch sizes due to its reliance on in-batch negatives, this objective is independent of cross-sample comparisons.

As shown in \cref{tab:und_ablation}, end-to-end training with the generative objective alone achieves an ImageNet linear-probe accuracy of $77.71$.
Adding latent-level SigLIP supervision improves the accuracy to $79.25$, while slightly improving generation and reconstruction, reaching $0.90$ GenEval, $83.76$ DPG-Bench, and $0.39$ rFID.
VLM-NLL supervision yields a substantially larger semantic improvement, increasing linear-probe accuracy to $81.69$, which exceeds the $79.64$ accuracy of the pretrained Qwen3-VL ViT.
Meanwhile, it largely preserves generative performance, achieving $0.89$ GenEval and $83.49$ DPG-Bench, with only a moderate change in reconstruction quality.
These results show that representation supervision can be incorporated into our end-to-end framework while largely preserving the generation-friendly properties of the learned latent space.
Overall, these results suggest that end-to-end latent learning can extend beyond generation and jointly support generation, representation learning, and reconstruction within a shared latent space.

\begin{table}[t]
\centering
\caption{
Text-to-image and image-to-text performance of frozen text/image VAEs versus our
end-to-end trained counterparts, all with the same block-causal MMDiT and the same
three-stage recipe. Image-to-text is evaluated on the COCO~\cite{lin2014microsoft} Karpathy test split with
pycocoevalcap~\cite{pycocoevalcap}, following~\cite{chameleon,transfusion}.
}
\label{tab:mm_e2e}
\setlength\tabcolsep{5pt}
\centering
\tiny
\begin{tabular}{lcccccccc}
\toprule
 \multirow{2}{*}{Setting} & \multicolumn{2}{c}{Text-to-image} & \multicolumn{6}{c}{Image-to-text} \\
 \cmidrule(lr){2-3} \cmidrule(lr){4-9}
 & GenEval$\uparrow$ & DPG-Bench$\uparrow$ & CIDEr$\uparrow$ & BLEU-4$\uparrow$ & METEOR$\uparrow$ & ROUGE-L$\uparrow$ & SPICE$\uparrow$ & CLIPScore$\uparrow$ \\
\midrule
Frozen text VAE + frozen image VAE & 79.88 & 63.96 & 20.03 & 5.36 & 13.72 & 29.01 & 8.03 & 53.21 \\
E2E-trained text VAE + frozen image VAE & 81.87 & 68.24 & 30.19 & 7.46 & 16.81 & 32.95 & 11.12 & 55.96 \\
E2E-joint-trained text VAE and image VAE & \textbf{86.57} & \textbf{72.68} & \textbf{33.28} & \textbf{7.76} & \textbf{17.74} & \textbf{34.16} & \textbf{11.81} & \textbf{57.97} \tabularnewline
\bottomrule
\end{tabular}
\end{table}

\begin{figure}[]
\centering
\includegraphics[width=\linewidth]{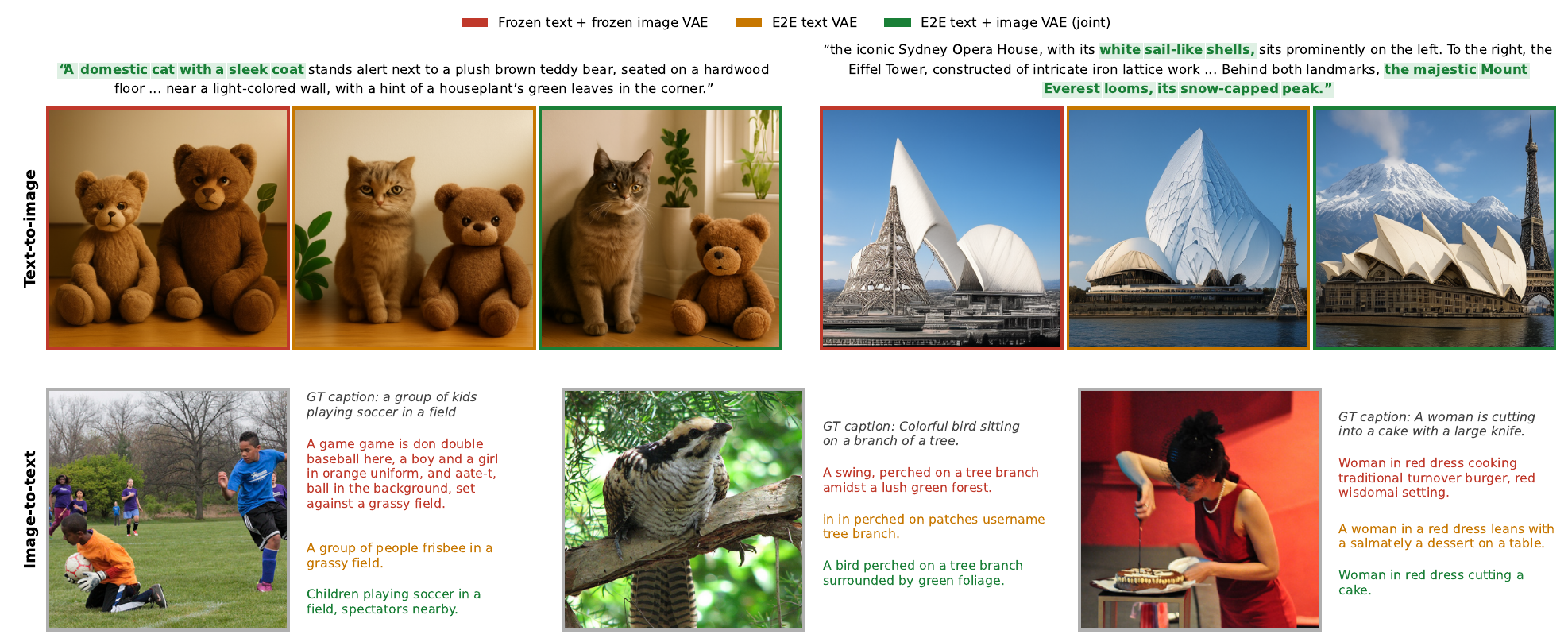}
\caption{
\textbf{Qualitative comparison of the three latent-space settings.}
All three settings use the same MMDiT and sampling configuration.
\textbf{Top:} text-to-image generation on DPG-Bench.
\textbf{Bottom:} image-to-text generation on COCO Karpathy test split.
End-to-end trained latents improve prompt fidelity and caption coherence, with joint text--image latent training producing the strongest qualitative results.
}
\label{fig:mm_qualitative}
\end{figure}

\textbf{Unified text-image modeling via end-to-end diffusion.}
We further investigate whether our end-to-end formulation can extend beyond image latents to jointly model text and images.
As shown in \cref{fig:further_exp}(c), we encode text into continuous latent representations with a text VAE and reconstruct the original text through a decoder using a cross-entropy (CE) loss.
We then feed both text and image latents into a block-causal MMDiT~\cite{mmdit} and train text-to-image $p(i|t)$ and image-to-text $p(t|i)$ generation.
The block-causal MMDiT uses the same model scale as the MMDiT-XL/2 in our text-to-image experiments.
The diffusion objectives from both directions shape the latent representations, while the entropy term prevents latent collapse.
We apply the same GenFirst strategy to balance latent modeling and reconstruction.
Due to computational constraints, the end-to-end phase uses only half as many training steps as our text-to-image experiment, while the subsequent prior-only phase uses only $30.8\%$ of its text-to-image training samples.

We initialize the text VAE from Cola~\cite{cola} and first replace the frozen text VAE with its end-to-end trained counterpart while keeping the image VAE frozen.
As shown in \cref{tab:mm_e2e}, this improves GenEval from $79.88$ to $81.87$ and DPG-Bench from $63.96$ to $68.24$, while also substantially improving image-to-text generation, with CIDEr increasing from $20.03$ to $30.19$.
Jointly training both the text and image VAEs further improves performance to $86.57$ on GenEval and $72.68$ on DPG-Bench, while achieving a CIDEr~\cite{vedantam2015cider} score of $33.28$ for image-to-text generation.
Consistent improvements are also observed across BLEU-4, METEOR, ROUGE-L, SPICE, and CLIPScore.
The qualitative comparisons in \cref{fig:mm_qualitative} show the same trend: end-to-end trained latents improve prompt fidelity in text-to-image generation and produce more coherent and accurate captions in image-to-text generation.
These results show that end-to-end latent learning extends beyond images and can jointly improve continuous text and image representations within a unified generative model.

\section{Conclusion}
We revisit end-to-end training of VAEs and generative models and make two key
observations. First, we find that latent collapse in naive end-to-end training is
mainly caused by a prior-entropy imbalance. 
Second, we identify asymmetric learning dynamics underlying the generation--reconstruction conflict.
Based on these observations, we introduce an entropy-preserving end-to-end objective that enables stable direct joint training and further propose GenFirst, a generation-before-reconstruction strategy that mitigates the generation--reconstruction conflict.
Experiments show that our method learns more generation-friendly
latent spaces and improves the generation--reconstruction trade-off. We hope this
work provides a simple and practical recipe for stable end-to-end latent
generative modeling.

\section{Limitations}
Although GenFirst enables stable end-to-end training and substantially alleviates the generation--reconstruction conflict, it does not completely eliminate the trade-off.
The end-to-end trained VAE therefore still sacrifices some reconstruction fidelity, as reflected by metrics such as PSNR.
Further improving reconstruction while preserving a generation-friendly latent space remains an important direction for future work.
Additionally, EAR is limited by the current CFG strategy for continuous autoregressive models.
Without CFG, scaling the EAR prior consistently improves gFID, whereas gFID with CFG does not improve accordingly.
Developing stronger guidance methods for continuous autoregressive models is therefore left for future work.
Finally, our text-to-image experiments use a substantially smaller training corpus than large-scale text-to-image models.
Scaling the data and training budget may further reveal the potential of end-to-end learned latent spaces.

\section*{Acknowledgments}
We would like to extend our deepest appreciation to Qinyu Zhao, Qiushan Guo, Shuai Wang, Shuchen Xue, Hongcan Guo, and Yian Zhao for their insightful discussions and valuable support.

\beginappendix

\section{More Analysis}

\textbf{Fine-grained evaluation of text-to-image generation.}
We further report the detailed GenEval and DPG-Bench results of our two text-to-image model variants in \cref{tab:t2i_eval_details}.
Version A performs the end-to-end phase on ImageNet, whereas Version B performs the end-to-end phase on the text-to-image training data.
The detailed training configurations of both variants are provided in \cref{app:training_details}.
Version A performs slightly better overall, while the two variants remain close, indicating that end-to-end VAE training is effective on both ImageNet and text-to-image training data.

\begin{table}[]
\centering
\scriptsize
\caption{
\textbf{Fine-grained evaluation of text-to-image generation.}
We report category-wise results on GenEval~\cite{geneval} and DPG-Bench~\cite{dpg} for two model variants.
}
\label{tab:t2i_eval_details}
\setlength{\tabcolsep}{4.0pt}

\begin{minipage}{\linewidth}
\centering
\textbf{GenEval}\\[2pt]
\begin{tabular}{lccccccc}
\toprule
Method
& Single Obj.
& Two Obj.
& Counting
& Color
& Position
& Color Attr.
& Overall \\
\midrule
EiT (Version A)
& \textbf{100.00}
& \textbf{94.44}
& \textbf{83.12}
& 94.15
& 86.75
& \textbf{81.75}
& \textbf{0.8987} \\
EiT (Version B)
& 99.06
& 93.18
& 82.81
& \textbf{94.95}
& \textbf{88.25}
& 78.00
& 0.8920 \\
\bottomrule
\end{tabular}
\end{minipage}

\vspace{5pt}

\begin{minipage}{\linewidth}
\centering
\textbf{DPG-Bench}\\[2pt]
\begin{tabular}{lcccccc}
\toprule
Method
& Global
& Entity
& Attribute
& Relation
& Other 
& Overall \\
\midrule
EiT (Version A)
& \textbf{88.30}
& 86.91
& \textbf{88.75}
& \textbf{89.64}
& 85.66 
& \textbf{82.60}\\
EiT (Version B)
& 84.26
& \textbf{88.10}
& 87.16
& 88.87
& \textbf{89.23} 
& 82.06\\
\bottomrule
\end{tabular}
\end{minipage}

\end{table}

\textbf{Effect of end-to-end training on VAE reconstruction.}
We evaluate the reconstruction quality of different VAEs in \cref{tab:vae_reconstruction}.
Compared with their corresponding pretrained VAEs, end-to-end trained VAEs show a moderate degradation in reconstruction metrics while maintaining strong reconstruction quality overall.
For the FLUX.1-dev VAE, end-to-end training changes rFID from $0.18$ to $0.23/0.21$ and PSNR from $31.59$ to $29.43/29.37$ for Variants A/B.
A similar trend is observed for SD-VAE and VA-VAE: the end-to-end trained variants retain comparable rFID, PSNR, SSIM, and LPIPS, although some reconstruction fidelity is sacrificed.
This result is consistent with the generation--reconstruction trade-off discussed in \cref{subsec:no_single_optimum}: end-to-end training reshapes the latent space toward generative modeling at the cost of a moderate decrease in reconstruction quality.

\begin{table}[t]
\centering
\scriptsize
\caption{
\textbf{VAE reconstruction evaluation on ImageNet-256.}
While our end-to-end training substantially improves generation quality, the resulting VAEs maintain competitive reconstruction performance with only a minor degradation.
}
\label{tab:vae_reconstruction}
\setlength{\tabcolsep}{5.0pt}

\begin{tabular}{lcccc}
\toprule
VAE
& rFID$\downarrow$
& PSNR$\uparrow$
& SSIM$\uparrow$
& LPIPS$\downarrow$ \\
\midrule
FLUX.1-dev VAE~\cite{flux1}
& 0.18
& 31.59
& 0.90
& 0.05 \\
E2E-trained VAE (Version A, Flux.1-dev VAE init.)
& 0.23
& 29.43
& 0.85
& 0.07 \\
E2E-trained VAE (Version B, Flux.1-dev VAE init.)
& 0.21
& 29.37
& 0.85
& 0.07 \\
\midrule
SD-VAE~\cite{ldm}
& 0.74
& 25.67
& 0.72
& 0.13 \\
E2E-trained VAE ($f8d4$, SD-VAE init.)
& 0.57
& 25.13
& 0.71
& 0.14 \\
\midrule
VA-VAE~\cite{vavae}
& 0.28
& 26.32
& 0.76
& 0.11 \\
E2E-trained VAE ($f16d32$, VA-VAE init.)
& 0.36
& 25.82
& 0.74
& 0.12 \\
\bottomrule
\end{tabular}
\end{table}

\textbf{Ablation on GenFirst prior weights for SiT.}
We first use $(\lambda_{\mathrm{prior}}^{(1)},\lambda_{\mathrm{prior}}^{(2)})=(1,0.1)$, which already achieves reasonable generation and reconstruction performance.
Reducing the prior weights to $(0.1,0.01)$ further improves both generation and reconstruction, with rFID decreasing from $0.61$ to $0.57$ and FID decreasing from $4.44$ to $3.57$.
This suggests that $(1,0.1)$ still applies excessive prior pressure and does not reach a favorable generation--reconstruction trade-off.
When the weights are further reduced to $(0.01,0.01)$, reconstruction continues to improve, but generation degrades.
We therefore use $(0.1,0.01)$ as the default setting and leave finer tuning of the two-stage prior weights to future work due to computational constraints.

\begin{table}[t]
\centering
\scriptsize
\caption{
\textbf{Ablation on GenFirst prior weights for SiT.}
We vary the prior weights in the generation-first and reconstruction-refinement stages.
}
\label{tab:sit_prior_weights}
\setlength{\tabcolsep}{4.5pt}

\begin{tabular}{cc|cccccc}
\toprule
$\lambda_{\mathrm{prior}}^{(1)}$
& $\lambda_{\mathrm{prior}}^{(2)}$
& rFID$\downarrow$
& PSNR$\uparrow$
& SSIM$\uparrow$
& LPIPS$\downarrow$
& FID$\downarrow$
& IS$\uparrow$ \\
\midrule
1.00
& 0.10
& 0.61
& 24.57
& 0.70
& 0.15
& 4.44
& 154.25 \\
0.10
& 0.01
& 0.57
& 25.13
& 0.71
& 0.14
& \textbf{3.57}
& \textbf{166.49} \\
0.01
& 0.01
& \textbf{0.48}
& \textbf{25.17}
& \textbf{0.72}
& \textbf{0.14}
& 3.84
& 161.40 \\
\bottomrule
\end{tabular}
\end{table}

\textbf{Effect of end-to-end training and REPA.}
We further disentangle the contributions of end-to-end training and REPA~\cite{repa} in \cref{tab:e2e_repa_ablation}.
Using SD-VAE~\cite{ldm} with vanilla SiT-XL/2~\cite{sit} yields an FID of $17.20$, while applying REPA during the prior-only phase improves it to $7.90$.
Replacing the frozen SD-VAE with our end-to-end trained VAE already improves the vanilla SiT-XL/2 result to an FID of $8.09$, even without REPA in either training phase.
Following REPA-E~\cite{repa_e}, introducing REPA during the end-to-end phase further improves the FID to 6.83.
Applying REPA again during the subsequent prior-only phase further reduces the FID to $3.57$.
These results show that end-to-end VAE training and REPA provide complementary rather than conflicting benefits: end-to-end training reshapes the latent space to make it easier for the generative model to learn, while REPA further improves generation through representation alignment.

\begin{table}[t]
\centering
\scriptsize
\caption{
\textbf{Effect of end-to-end training and REPA.}
We disentangle the contributions of end-to-end latent learning and representation alignment~\cite{repa} to SiT~\cite{sit} generation.
}
\label{tab:e2e_repa_ablation}
\setlength{\tabcolsep}{7.0pt}

\begin{tabular}{cccccc}
\toprule
\multicolumn{3}{c}{End-to-end phase} & \multicolumn{2}{c}{Prior-only phase (80 epochs)} & \multirow{2}{*}{FID$\downarrow$}\\
\cmidrule(lr){1-3} \cmidrule(lr){4-5}
VAE
& Model & REPA & Model & REPA
&  \\
\midrule
SD-VAE~\cite{ldm}
& - & - & SiT-XL/2~\cite{sit} & \xmark
& 17.20 \\
SD-VAE~\cite{ldm}
& - & - & SiT-XL/2~\cite{sit} & \cmark
& 7.90 \\
\midrule
E2E-trained VAE (SD-VAE init.) & SiT-XL/2~\cite{sit} & \xmark & SiT-XL/2~\cite{sit} & \xmark
& 8.09 \\
E2E-trained VAE (SD-VAE init.) & SiT-XL/2~\cite{sit} & \cmark & SiT-XL/2~\cite{sit} & \xmark
& 6.83 \\
E2E-trained VAE (SD-VAE init.) & SiT-XL/2~\cite{sit} & \cmark & SiT-XL/2~\cite{sit} & \cmark
& \textbf{3.57} \\
\bottomrule
\end{tabular}
\end{table}

\section{A Shortcut Solution in Naive End-to-End Training}
\label{app:collapse_shortcut}

When the learned prior objective is strong while posterior entropy is weak, we
observe that the encoder can reduce both the posterior variance and the diversity
of posterior means. In the extreme case,
\[
\sigma_\phi^2(x)\rightarrow 0,
\qquad
\mu_\phi(x)\approx c,
\]
where $c$ is nearly constant across different images.
Since the posterior noise becomes very small, the decoder may still exploit
small residual differences in $\mu_\phi(x)$ for reconstruction, even though the
overall latent distribution is already highly concentrated.

For a diffusion or flow-matching prior, such a concentrated latent distribution
also makes the denoising objective artificially easy. Consider the linear
interpolation used by SiT,
\[
z_t=(1-t)\epsilon_t+t z,
\qquad
v=z-\epsilon_t.
\]
If the latent collapses to $z\approx c$, then
\[
z_t\approx(1-t)\epsilon_t+t c,
\qquad
v\approx\frac{c-z_t}{1-t}.
\]
Therefore, predicting the target velocity requires little knowledge of the image
distribution: the model can approximately recover the target through a simple
affine transformation of the noisy input around the nearly constant code $c$.
The denoising loss can thus become small even though the generative model has
not learned a meaningful latent distribution.

This explains why preserving posterior entropy is important during end-to-end
training. Increasing the entropy prevents $\sigma_\phi^2(x)$ from vanishing.
Under non-trivial posterior noise, if all posterior means were also nearly
identical, the latent samples of different images would become indistinguishable
to the decoder. The reconstruction objective therefore encourages
$\mu_\phi(x)$ to remain image-dependent and distributed across the latent space.
Together, entropy preservation and reconstruction discourage both variance and
mean collapse.
Motivated by this observation, we tie the entropy weight to the prior weight in
our default end-to-end objective,
$\lambda_{\mathrm{ent}}=\lambda_{\mathrm{prior}}$.
This keeps posterior-entropy preservation coupled to the strength of prior-fitting pressure and avoids the degenerate shortcut observed in naive end-to-end training.

\section{Implementation Details}
\label{app:implementation_details}

\subsection{Training and Inference Details}
\label{app:training_details}
We provide the complete architecture, training, and inference configurations for
EAR, EiT, and text-to-image EiT in
\cref{tab:ear_details,tab:eit_details,tab:t2i_details}.
Below, we summarize the training pipeline and the relationship between different
resolutions and model variants.

\textbf{EAR on ImageNet.}
The detailed EAR configuration is provided in \cref{tab:ear_details}.
For ImageNet $256\times256$, we jointly train the MAR VAE~\cite{mar} and the continuous AR prior~\cite{zheng2025farmer} from scratch using the two-stage GenFirst schedule.
The resulting VAE is then reused for the resolution $512\times512$ experiment: we keep the
VAE frozen and train a larger EAR prior on the higher-resolution latent
sequences.
Therefore, the resolution $512\times512$ experiment evaluates whether the latent space
learned by GenFirst at resolution $256\times256$ transfers to higher-resolution generative
modeling without further adapting the tokenizer.
We follow FARMER~\cite{zheng2025farmer} for the continuous-AR parameterization
and resampling-based classifier-free guidance.

\textbf{EiT on ImageNet.}
The ImageNet EiT configuration is summarized in \cref{tab:eit_details}.
At ImageNet $256\times256$, training consists of an end-to-end phase followed by a prior-only phase. 
During the end-to-end phase, the VAE and SiT are jointly optimized using the two-stage GenFirst schedule. 
We then freeze the resulting VAE and train a new SiT on its latent space. 
For resolution $512\times512$, we directly reuse and freeze the VA-VAE obtained from the resolution $256\times256$ end-to-end stage, and train only a new SiT at the higher resolution.
We observe that the latent space changes rapidly during end-to-end training, which can leave the decoder insufficiently adapted to the updated latents.
Therefore, for the ImageNet-$256$ EiT results in \cref{tab:imagenet256_sota} only, we additionally fine-tune the VAE decoder for 10K steps after the end-to-end stage. This improves the 80-epoch unguided gFID from $3.00$ to $2.79$. 
This decoder fine-tuning is not used in the other EiT experiments.
For the corresponding 80-epoch guided result, we use CFG scale $2.0$ with a guidance interval of $[0.2,0.85]$.
Unless otherwise specified, the optimization and REPA settings follow REPA-E~\cite{repa_e}; complete architecture, training, and inference configurations are provided in \cref{tab:eit_details}.

\textbf{Text-to-image EiT.}
We consider two variants for obtaining an end-to-end trained FLUX.1-dev VAE~\cite{flux1}, as detailed in \cref{tab:t2i_details}.
In \textit{Version A}, the VAE is jointly trained with a smaller MMDiT (MMDiT-L/2) directly on the text-to-image training data using GenFirst.
In \textit{Version B}, the VAE is jointly trained with SiT-XL/2 on ImageNet using the same GenFirst setting as class-conditional EiT.
These variants compare a VAE shaped directly by text-conditioned generation with one shaped by class-conditional ImageNet generation.
After the end-to-end phase, we freeze the resulting VAE and train the same MMDiT-XL/2 prior for both variants.
We first train at $256\times256$ for 200K steps, continue at $512\times512$ for 80K steps, and finally fine-tune on the OpenAI-4o dataset~\cite{open4o} at $512\times512$ for 40K steps.
Complete architecture, optimization, and inference settings are provided in \cref{tab:t2i_details}.

\begin{table*}[t]
\centering
\caption{
\textbf{Detailed training configuration of EAR on ImageNet.}
EAR-256 jointly trains the VAE and AR prior with GenFirst.
For EAR-512, we reuse and freeze the VAE obtained from EAR-256 and train only
the AR prior at the higher resolution.
}
\label{tab:ear_details}
\scriptsize
\setlength{\tabcolsep}{5pt}
\renewcommand{\arraystretch}{1.10}

\begin{tabular}{lcc}
\toprule
\textbf{Configuration}
& \textbf{EAR-256}
& \textbf{EAR-512} \\
\midrule

\multicolumn{3}{l}{\textit{Architecture}} \\
\midrule
VAE architecture
& MAR VAE~\cite{mar}
& EAR-256 VAE (frozen) \\

Latent configuration
& $f16d16$
& $f16d16$ \\

AR backbone
& EAR-312M
& EAR-940M \\

Transformer depth
& 24
& 36 \\

Hidden dimension
& 1024
& 1280 \\

Attention heads
& 16
& 20 \\

GMM components $K$
& 64
& 1024 \\

\midrule
\multicolumn{3}{l}{\textit{Training}} \\
\midrule

Trainable modules
& VAE + AR prior
& AR prior only \\

Training resolution
& $256\times256$
& $512\times512$ \\

Generation-first stage
& 500 epochs
& -- \\

$\lambda_{\mathrm{prior}} / \lambda_{\mathrm{ent}}$
& $1.0 / 1.0$
& -- \\

Reconstruction-refinement stage
& 140 epochs
& -- \\

$\lambda_{\mathrm{prior}} / \lambda_{\mathrm{ent}}$
& $0.25 / 0.25$
& -- \\

Prior-only training
& --
& 320 epochs \\

Global batch size
& 512
& 512 \\

Optimizer
& AdamW~\cite{adamw}, $\beta_1,\beta_2=0.9,0.999$
& AdamW~\cite{adamw}, $\beta_1,\beta_2=0.9,0.999$ \\

Learning rate
& $1\times10^{-4}$
& $1\times10^{-4}$ \\

LR schedule
& 10-epoch linear warmup, then constant
& 10-epoch linear warmup, then constant \\

EMA rate & 0.9999 & 0.9999 \\

Class token drop (for CFG) & 0.1 & 0.1 \\
\midrule
\multicolumn{3}{l}{\textit{Inference}} \\
\midrule

Eval batch size & 50 & 50 \\
Guidance
& resampling-based CFG~\cite{zheng2025farmer}
& resampling-based CFG~\cite{zheng2025farmer} \\

Weight temperature (Propose stage) & 0.9 & 0.9\\
Variance temperature (Propose stage) & 0.9 & 0.9\\
Sample number from $p_c/p_u$ & 4/4 & 4/4\\
Weight temperature (Weigh stage) & -0.55 & -0.5\\
Variance temperature (Weigh stage) & 1.1 & 1.0 \\
Resampling temperature & 0.9 & 1.0 \\
CFG scale
& 1.5
& 2.0 \\

\bottomrule
\end{tabular}
\end{table*}

\begin{table*}[t]
\centering
\caption{
\textbf{Detailed training configuration of EiT on ImageNet.}
EiT-256 first jointly trains the VAE and SiT with the two-stage GenFirst
schedule, after which the VAE is frozen and SiT training continues.
EiT-512 directly reuses and freezes the EiT-256 VAE and trains only SiT.
}
\label{tab:eit_details}
\scriptsize
\setlength{\tabcolsep}{5pt}
\renewcommand{\arraystretch}{1.10}

\begin{tabular}{lcc}
\toprule
\textbf{Configuration}
& \textbf{EiT-256}
& \textbf{EiT-512} \\
\midrule

\multicolumn{3}{l}{\textit{Architecture}} \\
\midrule

VAE
& SD-VAE / VA-VAE
& EiT-256 VAE (frozen, VA-VAE version) \\

Latent configuration
& $f8d4$/$f16d32$
& $f16d32$ \\

Generative backbone
& SiT-XL/2~\cite{sit} / SiT-XL/1~\cite{sit}
& SiT-XL/1~\cite{sit} \\

Transformer depth
& 28
& 28 \\

Hidden dimension
& 1152
& 1152 \\

Attention heads
& 16
& 16 \\

\midrule
\multicolumn{3}{l}{\textit{End-to-end phase}} \\
\midrule

Trainable modules
& VAE + SiT
& -- \\

Generation-first stage
& 80 epochs
& -- \\

$\lambda_{\mathrm{prior}} / \lambda_{\mathrm{ent}}$
& $0.1 / 0.1$
& -- \\

Reconstruction-refinement stage
& 40 epochs
& -- \\

$\lambda_{\mathrm{prior}} / \lambda_{\mathrm{ent}}$
& $0.01 / 0.01$
& -- \\

\midrule
\multicolumn{3}{l}{\textit{Prior-only phase}} \\
\midrule

VAE
& frozen
& frozen \\

Trainable modules
& SiT only
& SiT only \\

Training resolution
& $256\times256$
& $512\times512$ \\

Training duration
& 800 epochs
& 400 epochs \\

\midrule
\multicolumn{3}{l}{\textit{Shared optimization}} \\
\midrule

Global batch size
& 256
& 256 \\

Optimizer
& AdamW~\cite{adamw}, $\beta_1,\beta_2=0.9,0.999$
& AdamW~\cite{adamw}, $\beta_1,\beta_2=0.9,0.999$ \\

Learning rate
& $1\times10^{-4}$
& $1\times10^{-4}$ \\

LR schedule
& constant
& constant \\

EMA rate & 0.9999 & 0.9999 \\

Class token drop (for CFG) & 0.1 & 0.1 \\

\midrule
\multicolumn{3}{l}{\textit{REPA}} \\
\midrule
$\lambda$ for SiT & 0.5 & 0.5 \\
$\lambda$ for VAE & 1.5 & 1.5 \\
Alignment depth & 8 & 8 \\
Encoder & DINOv2-B~\cite{dino} & DINOv2-B~\cite{dino} \\
\midrule
\multicolumn{3}{l}{\textit{Inference}} \\
\midrule
Guidance 
& CFG~\cite{cfg}+Autoguidance~\cite{karras2024guiding}
& CFG \\

Sampler
& SDE
& SDE \\

Sampling steps 
& 250
& 250 \\

Weak model 
& 20 epochs
& -- \\

Autoguidance scale 
& 1.3 
& -- \\

CFG scale
& 1.5
& 2.5 \\

Guidance interval 
& [0.2, 0.7]
& [0.2, 0.8]\\

\bottomrule
\end{tabular}
\end{table*}

\begin{table*}[t]
\centering
\caption{
\textbf{Detailed training configuration of EiT for text-to-image generation.}
We consider two settings for the end-to-end latent-shaping phase:
joint training with SiT on ImageNet or with a smaller MMDiT on the T2I training
mixture. The resulting VAE is then frozen and used to train the target MMDiT.
}
\label{tab:t2i_details}
\scriptsize
\setlength{\tabcolsep}{5pt}
\renewcommand{\arraystretch}{1.10}

\begin{tabular}{lcc}
\toprule
\textbf{Configuration}
& \textbf{EiT (Version A)}
& \textbf{EiT (Version B)} \\
\midrule

\multicolumn{3}{l}{\textit{Architecture}} \\
\midrule

VAE
& FLUX.1-dev VAE~\cite{flux1}
& FLUX.1-dev VAE~\cite{flux1} \\

Latent configuration
& $f8d16$
& $f8d16$ \\

Text Encoder 
& --
& clip-vit-large-patch14~\cite{clip}\\

Generative backbone (E2E phase)
& SiT-XL/2~\cite{sit}
& MMDiT-L/2~\cite{mmdit} \\

Backbone depth (E2E phase)
& 28
& 24 \\

Hidden dimension (E2E phase)
& 1152
& 1024 \\

Attention heads (E2E phase)
& 16
& 16 \\

Training data (E2E phase)
& ImageNet
& T2I mixture \\

Training resolution (E2E phase)
& $256\times256$
& $256\times256$ \\
\midrule
Text Encoder 
& \multicolumn{2}{c}{clip-vit-large-patch14~\cite{clip}}\\

Generative backbone (Frozen VAE phase)
& \multicolumn{2}{c}{MMDiT-XL/2} \\

Backbone depth (Frozen VAE phase)
& \multicolumn{2}{c}{28} \\

Hidden dimension (Frozen VAE phase)
& \multicolumn{2}{c}{1152} \\

Attention heads (Frozen VAE phase)
& \multicolumn{2}{c}{16} \\

Training data (Frozen VAE phase)
& \multicolumn{2}{c}{T2I mixture} \\

Training resolution (Frozen VAE phase)
& \multicolumn{2}{c}{$256\times256$ (stage1), $512\times512$ (stage2, stage3)}\\

\midrule
\multicolumn{3}{l}{\textit{End-to-end phase}} \\
\midrule

Trainable modules
& VAE + SiT
& VAE + MMDiT \\

Global batch size 
& 256 
& 256 \\

Generation-first stage
& 80 epochs
& 400K steps \\

$\lambda_{\mathrm{prior}} / \lambda_{\mathrm{ent}}$
& $0.1 / 0.1$
& $0.1 / 0.1$ \\

Reconstruction-refinement stage
& 40 epochs
& 200K steps \\

$\lambda_{\mathrm{prior}} / \lambda_{\mathrm{ent}}$
& $0.01 / 0.01$
& $0.01 / 0.01$ \\

\midrule
\multicolumn{3}{l}{\textit{Frozen-VAE MMDiT training}} \\
\midrule

VAE
& \multicolumn{2}{c}{Frozen E2E-trained VAE} \\

T2I mixture, $256^2$
& \multicolumn{2}{c}{200K steps, global batch size 1536} \\

T2I mixture, $512^2$
& \multicolumn{2}{c}{80K steps, global batch size 512} \\

4o data fine-tuning, $512^2$
& \multicolumn{2}{c}{40K steps, global batch size 512} \\

\midrule
\multicolumn{3}{l}{\textit{Inference}} \\
\midrule

Sampler
& \multicolumn{2}{c}{Euler} \\

Sampling steps
& \multicolumn{2}{c}{50} \\

CFG scale
& \multicolumn{2}{c}{7.0} \\

CFG interval
& \multicolumn{2}{c}{[0.35, 1.0]} \\

\bottomrule
\end{tabular}
\end{table*}

\subsection{Details of Loss-Balancing Baselines}
\label{app:loss_balancing}

We provide additional implementation details for the loss-balancing baselines
compared in \cref{tab:ablation_pz_method}. All variants use the same EAR
architecture, dataset, and reconstruction-related loss weights; they differ only
in how the influence of the generative objective is controlled during training.

\textbf{Constant weighting.}
We use a fixed prior weight $\lambda_{\mathrm{prior}}=1$ throughout 320 training epochs, with
$\lambda_{\mathrm{ent}}=\lambda_{\mathrm{prior}}$.
We observe that training largely converges around 260 epochs, with little additional improvement from longer training.
Moreover, as shown in \cref{fig:abl_1}(a), smaller constant prior weights consistently yield worse generation--reconstruction trade-offs than $\lambda_{\mathrm{prior}}=1$.
We therefore report $\lambda_{\mathrm{prior}}=1$ as the strongest constant-weight baseline.

\textbf{Cosine decay.}
We train for 320 epochs and smoothly decay the prior weight from $0.5$ to $0.1$
with a cosine schedule. The entropy weight follows the same schedule,
i.e., $\lambda_{\mathrm{ent}}=\lambda_{\mathrm{prior}}$. We performed a simple hyperparameter sweep but did not obtain better results than the constant-weight setting.

\textbf{PI-based adaptive control.}
We additionally test a feedback-based strategy inspired by ControlVAE~\cite{shao2020controlvae}. 
Instead of prescribing a fixed schedule, the controller adjusts how strongly the generative objective updates the VAE encoder through a $w$. 
Thus, $w=1.0$ allows the full generative gradient to propagate to the encoder, while smaller $w$ weakens this update.
The controller uses the per-step reconstruction error as feedback with target $\ell^{*}=0.055$ following ControlVAE~\cite{shao2020controlvae}. 
We initialize $w=1.0$ and constrain it to $[0.004096,1.0]$. 
In our experiments, the reconstruction error is substantially above the target during early training, causing $w$ to rapidly decrease toward its lower bound. 
It subsequently oscillates between different values, leading to abrupt changes in the gradient received by the latent encoder and numerical instability. 
We also performed simple tuning of the controller hyperparameters, but did not obtain stable training under the tested configurations.

\subsection{Toward a Shared Latent Space for Generation and Representation Learning}
\label{app:und}

As discussed in \cref{exp:further_extensions}, we further examine whether the
end-to-end framework can learn a visual representation that simultaneously
supports reconstruction, generation, and representation learning. In this section, we
describe the autoencoder architecture, training schedule, and discrimination
objectives used in the three settings reported in
\cref{tab:und_ablation}.

\textbf{Qwen3-VL-based visual tokenizer.}
We replace the convolutional encoder of the FLUX.1-dev VAE~\cite{flux1} with the vision tower of Qwen3-VL-2B-Instruct~\cite{bai2025qwen3}. 
At an input resolution of \(256\times256\), the patch size of 16 produces a \(16\times16\) grid of visual tokens. Each token is represented by a 1,024-dimensional feature before the spatial merger.
We construct a Gaussian projector on top of these features. 
We first apply a LayerNorm initialized from the normalization layer of the pretrained Qwen3-VL merger. 
Two independent linear projections then map each 1,024-dimensional visual token to 64 dimensions, predicting the posterior mean and log-variance, respectively:
\begin{equation*}
    \boldsymbol{\mu}_i =
    W_{\mu}\operatorname{LN}(\mathbf{h}_i)+b_{\mu},
    \qquad
    \log\boldsymbol{\sigma}^{2}_i =
    W_{\sigma}\operatorname{LN}(\mathbf{h}_i)+b_{\sigma},
\end{equation*}
where \(\mathbf{h}_i\in\mathbb{R}^{1024}\) and
\(\boldsymbol{\mu}_i,\log\boldsymbol{\sigma}^{2}_i\in\mathbb{R}^{64}\).
The 64 outputs associated with each ViT token are reshaped into a \(2\times2\) block with 16 channels. 
The resulting latent therefore has shape \(16\times32\times32\), matching both the dimensionality and spatial rate of the FLUX \(f8d16\) latent. 
This allows us to retain the generative architecture and training setting used in our text-to-image experiments.
The decoder is randomly initialized and consists of a 12-layer ViT with width 1,024 and 16 attention heads. 
Each decoder token predicts a \(16\times16\) RGB patch. 
We additionally employ two lightweight \(3\times3\) convolutional layers after assembling the image to reduce visible artifacts at patch boundaries.

\textbf{Decoder warm-up.}
Directly optimizing the randomly initialized decoder together with the pretrained vision tower exposes the encoder to large and uninformative reconstruction gradients at the beginning of training. 
In preliminary experiments, this rapidly degraded the semantic structure inherited from Qwen3-VL. 
We therefore first freeze the Qwen3-VL vision tower and train only the Gaussian bottleneck and decoder for 100K steps.
In this warm-up stage, we apply only a weak KL regularizer with weight \(10^{-6}\); the adversarial loss is enabled after 10K steps. 

\textbf{End-to-end training.}
After decoder warm-up, we unfreeze the vision tower and jointly optimize the encoder, projector, decoder, and an MMDiT-L/2 generative prior. Apart from the discrimination objective, this stage follows the text-to-image end-to-end training recipe used in our main experiments, including the reconstruction losses, flow-matching objective, entropy regularization, and REPA alignment to DINOv2 features.

Due to the available computational budget, we shorten the original two-stage schedule by a factor of two. 
We first train for 200K steps with the prior weight set to \(0.1\).
We then train for another 100K steps with the prior weight set to \(0.01\).
For the two discrimination variants, the discrimination-loss weight is the same as the prior weight. 
The global batch size is 256. 
We compare the following three settings.

\textbf{Stable E2E without discrimination supervision.}
The first setting uses only reconstruction, generative, entropy, and REPA
objectives.
It serves as the generation--reconstruction baseline and does not receive an
explicit discrimination objective.

\textbf{Latent-level SigLIP supervision.}
In the second setting, discrimination supervision is applied directly to the same normalized latent consumed by MMDiT, which is a \(16\times16\) sequence of 64-dimensional tokens. 
A linear stem maps each token to width 1,024, after which a learnable query performs attention pooling. 
The pooled feature is projected to the 768-dimensional CLIP image--text embedding space.
Given normalized image and text embeddings \(\mathbf{v}_i\) and \(\mathbf{t}_j\), we use the sigmoid pairwise objective~\cite{zhai2023sigmoid}
\begin{equation}
    \mathcal{L}_{\mathrm{Sig}}
    =
    -\frac{1}{B}
    \sum_{i=1}^{B}\sum_{j=1}^{B}
    \log \sigma
    \left[
        y_{ij}
        \left(
            \tau\,\mathbf{v}_i^\top\mathbf{t}_j+\beta
        \right)
    \right],
\end{equation}
where \(y_{ij}=1\) for a matched image--caption pair and \(-1\) otherwise.
Text features and negatives are gathered across workers, so the loss uses the global batch.

We complement this objective with classification-based caption supervision~\cite{huang2024classification}.
The pooled image feature predicts a bag of CLIP caption tokens. 
Caption tokens form a multi-hot target, which is reweighted using online inverse document frequency and normalized into a probability distribution. 
The corresponding soft cross-entropy is
\begin{equation}
    \mathcal{L}_{\mathrm{CLS}}
    =
    -\sum_{k}
    \widetilde{q}_{k}
    \log p_{k},
    \qquad
    \widetilde{q}_{k}
    \propto
    q_k\log\frac{N}{f_k}.
\end{equation}
This weighting suppresses frequent syntactic tokens and concentrates the gradient on visually informative content words. 
The complete discrimination objective is
\begin{equation}
    \mathcal{L}_{\mathrm{disc}}^{\mathrm{SigLIP}}
    =
    \mathcal{L}_{\mathrm{Sig}}
    + \mathcal{L}_{\mathrm{CLS}}.
\end{equation}
This objective directly shapes the latent representation used by the generative model.

\textbf{Trunk-level VLM-NLL supervision.}
In the third setting, we preserve the interface between the Qwen3-VL vision tower and its pretrained language model. 
Rather than applying supervision after the 16-channel bottleneck latent, we branch from the ViT trunk and feed its native merged visual tokens to the frozen Qwen3-VL-2B language model. 
We also retain the original DeepStack features injected into the early language-model layers.
For an image \(x\) and its caption \(\mathbf{c}=(c_1,\ldots,c_T)\), the discrimination loss is the caption-token negative log-likelihood
\begin{equation}
    \mathcal{L}_{\mathrm{disc}}^{\mathrm{NLL}}
    =
    -\frac{1}{T}
    \sum_{t=1}^{T}
    \log p_{\mathrm{Qwen}}
    \left(
        c_t
        \mid c_{<t},\,f_{\mathrm{ViT}}(x)
    \right).
\end{equation}
The Qwen3-VL language model, language-model head, spatial merger, and DeepStack mergers are all frozen. 
Consequently, the NLL can only be reduced by preserving visual features that remain readable by the original Qwen3-VL language model. 
Unlike SigLIP supervision, the VLM-NLL objective does not depend on cross-sample negatives and therefore does not require a large global batch.

\textbf{Evaluation.}
We evaluate reconstruction using posterior-mean reconstructions on 50K
ImageNet validation images and report rFID and PSNR. Semantic accessibility is
measured by an ImageNet ridge linear probe trained on 100K images and evaluated
on 10K validation images. 
For final generation evaluation, we freeze each trained visual tokenizer and train the same MMDiT-XL pipeline used in our text-to-image experiments before evaluating GenEval and DPG-Bench.

\subsection{Unified Text--Image Modeling via End-to-End Diffusion}
\label{app:mm}

As discussed in \cref{exp:further_extensions}, our end-to-end formulation is not restricted to image latents and can be applied to other modalities.
We therefore extend our framework to text and jointly model text and images within a single diffusion backbone, allowing the generative objective to shape both text and image latent spaces.

\textbf{Text latent representation.}
We initialize the text VAE from Cola~\cite{cola}.
Each text token is encoded into a $16$-dimensional continuous latent, and the decoder reconstructs the original tokens with a cross-entropy loss.
We train the encoder and decoder trunks together with the latent projections ($\sim$194M trainable parameters), while freezing the token embedding and output vocabulary head.
The encoder applies a non-affine LayerNorm to the posterior mean, so we do not apply the additional latent normalization used for image latents.
To handle variable-length captions efficiently, text sequences are packed without padding and converted to a dense $(B,L,d)$ representation with a validity mask only before entering the diffusion model.

\textbf{Block-causal joint diffusion.}
During the end-to-end phase, we jointly model text and image latents using a block-causal MMDiT-L/2~\cite{mmdit} with $24$ blocks, width $1024$, $16$ attention heads, and patch size $2$.
The model maintains separate streams for the two modalities and couples them through joint attention.
At $256\times256$ resolution, each image contains $16\times16=256$ latent tokens, while text contains at most $256$ latent tokens and is partitioned into blocks of $16$ tokens.
Image tokens use bidirectional attention, whereas text tokens use block-causal attention, enabling block-wise generation of $p(t|i)$.

We train text-to-image $p(i|t)$ and image-to-text $p(t|i)$ with equal sampling probability.
For each sample, only the target modality is noised and supervised by the diffusion objective, while the conditioning modality is kept clean at $t=0$.
Image timesteps are sampled independently per example, whereas text timesteps are sampled independently for each text block.
For text generation, each noisy block attends to all preceding clean blocks and to the noisy tokens within the current block, allowing all blocks to be trained in parallel with clean prefix conditioning.
Cross-modal attention is direction-dependent: image tokens attend to clean text for $p(i|t)$, while text tokens attend to clean image features for $p(t|i)$.
For classifier-free guidance, we drop the conditioning modality with probability $0.1$ by disabling the corresponding cross-modal attention.

\textbf{Training objective.}
The text VAE follows the same end-to-end formulation as the image branch:
\begin{equation}
\mathcal{L}_{\mathrm{text}}
=
\mathcal{L}_{\mathrm{CE}}
+
\lambda_{\mathrm{KL}}
D_{\mathrm{KL}}\!\left(q_{\phi_t}(z_t|x)\,\|\,\mathcal{N}(0,I)\right)
-
\lambda_{\mathrm{ent}}
\mathcal{H}\!\left(q_{\phi_t}(z|x)\right)
+
\lambda_{\mathrm{prior}}
\mathcal{L}_{\mathrm{prior}},
\end{equation}
where $\mathcal{L}_{\mathrm{prior}}$ is the flow-matching objective of the joint diffusion model and $\lambda_{\mathrm{KL}}=10^{-6}$.
Following our analysis in \cref{subsec:prior_entropy}, the entropy term counteracts the prior-fitting pressure and prevents latent collapse.
We apply the entropy term to a modality only when it is generated, since a clean conditioning modality does not receive prior-fitting pressure from the diffusion objective.
For text-only end-to-end training, the image VAE remains frozen and only the text encoder receives generative gradients.
For joint end-to-end training, both text and image VAEs are optimized under the same GenFirst schedule, allowing the bidirectional diffusion objectives to shape both latent spaces.

\textbf{Training and evaluation.}
We compare three latent-training settings.
In the first setting, both the Cola text VAE and the FLUX.1-dev image VAE~\cite{flux1} are kept frozen, providing a fully frozen latent baseline.
In the second setting, we end-to-end train the text VAE together with the block-causal MMDiT while keeping the FLUX.1-dev image VAE frozen, isolating the effect of reshaping the text latent space.
In the third setting, both the Cola-initialized text VAE and the FLUX.1-dev image VAE are jointly optimized with the block-causal MMDiT, allowing the bidirectional generative objectives to reshape both latent spaces.
End-to-end training is conducted at $256\times256$ resolution on the same dataset as our text-to-image experiments, with captions truncated to $256$ tokens.
We use a batch size of $256$, AdamW~\cite{adamw} with learning rates of $10^{-4}$ for MMDiT and $10^{-5}$ for the text VAE, and gradient clipping at $1.0$.
We use $\lambda_{\mathrm{prior}}=\lambda_{\mathrm{ent}}=0.1$ for the first $200$K steps and reduce both to $0.01$ for another $100$K steps.
For joint end-to-end training, the same GenFirst schedule is applied to both text and image latent optimization.

After end-to-end latent training, we freeze the resulting text and image VAEs and train a new block-causal MMDiT-XL/2 from scratch.
The prior-only model uses $28$ blocks, width $1152$, and $16$ attention heads, matching the model scale of MMDiT-XL/2 in our main text-to-image experiments.
We follow the same training protocol as our text-to-image experiments: $200$K steps at $256\times256$, $80$K steps at $512\times512$, followed by $40$K steps of fine-tuning on GPT-4o data.
We evaluate text-to-image generation using GenEval and DPG-Bench. We evaluate image-to-text generation on the COCO Karpathy test split using CIDEr, BLEU-4, METEOR, ROUGE-L, SPICE, and CLIPScore.

\section{More Qualitative Results}
\begin{figure}
    \centering
    \includegraphics[width=\linewidth]{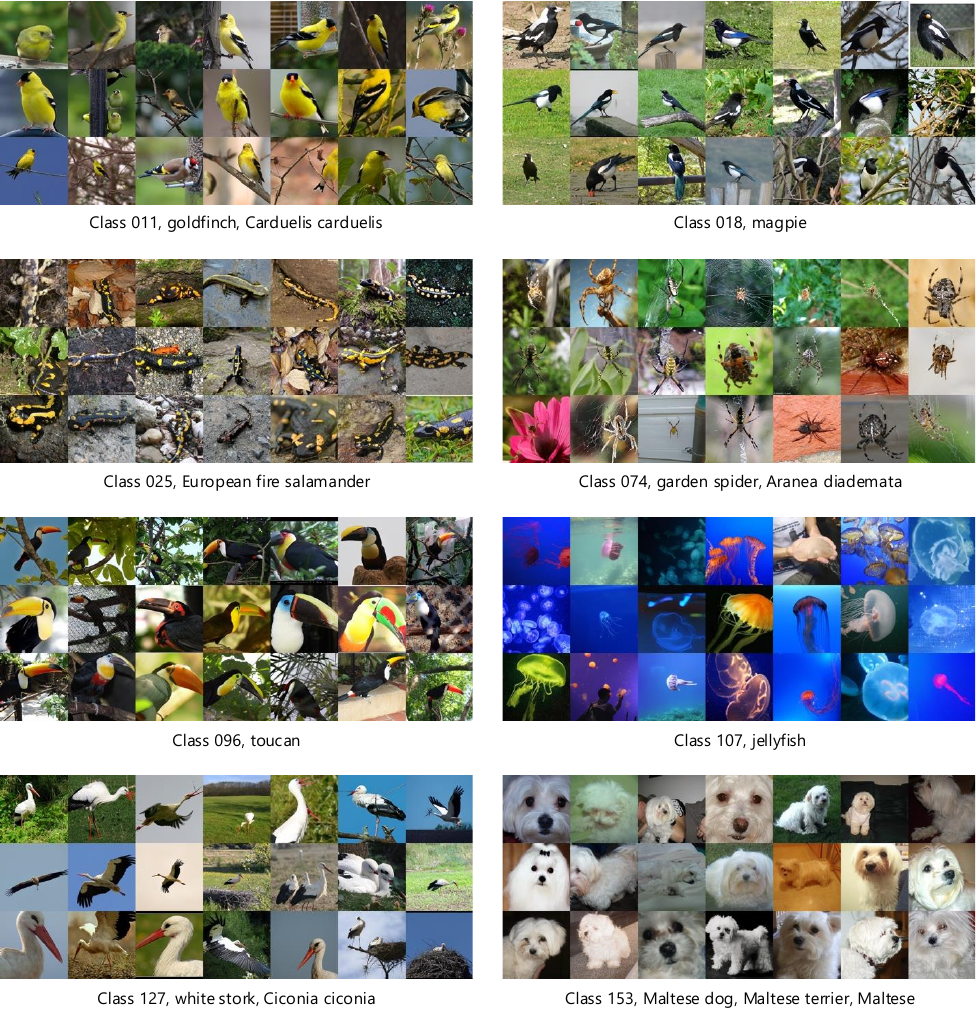}
    \caption{\textbf{Uncurated samples from EiT (480 epochs) with guidance on ImageNet $256\times256$.}}
    \label{fig:vis_dm1}
\end{figure}

\begin{figure}
    \centering
    \includegraphics[width=\linewidth]{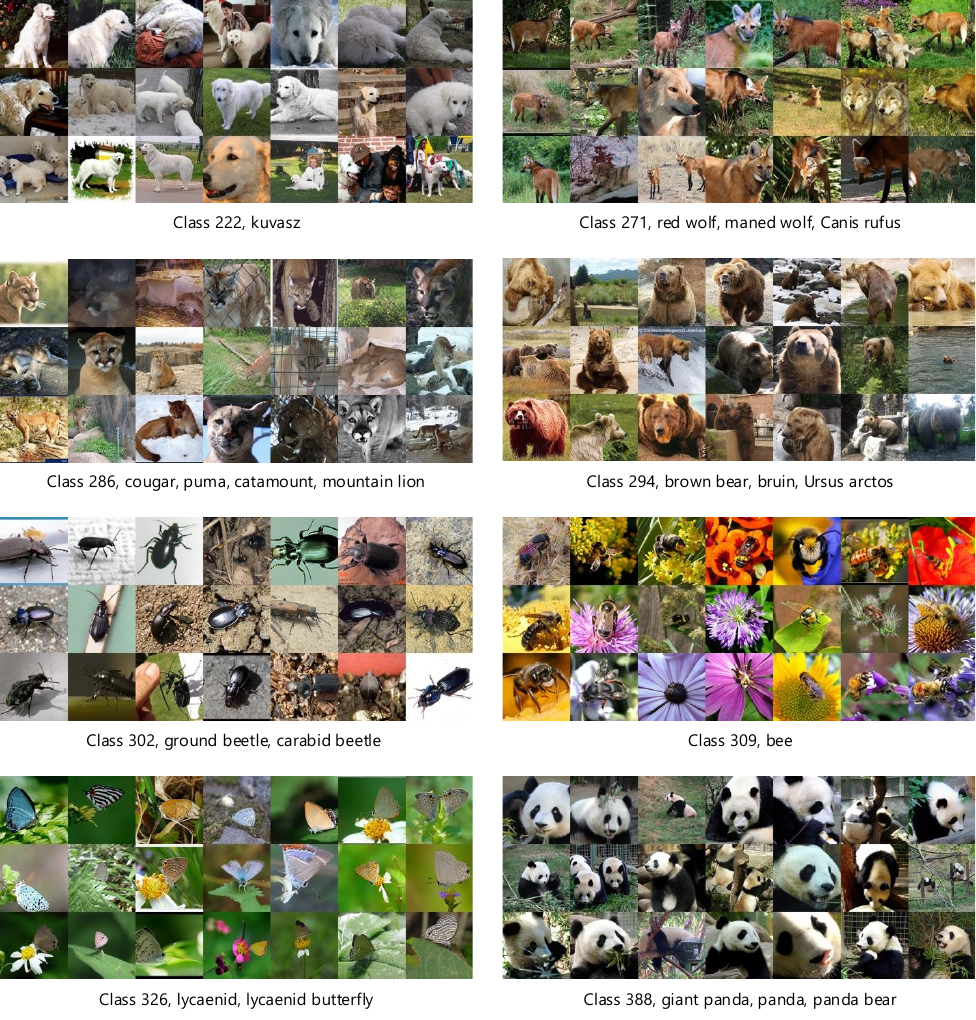}
    \caption{\textbf{Uncurated samples from EiT (480 epochs) with guidance on ImageNet $256\times256$.}}
    \label{fig:vis_dm2}
\end{figure}

\begin{figure}
    \centering
    \includegraphics[width=\linewidth]{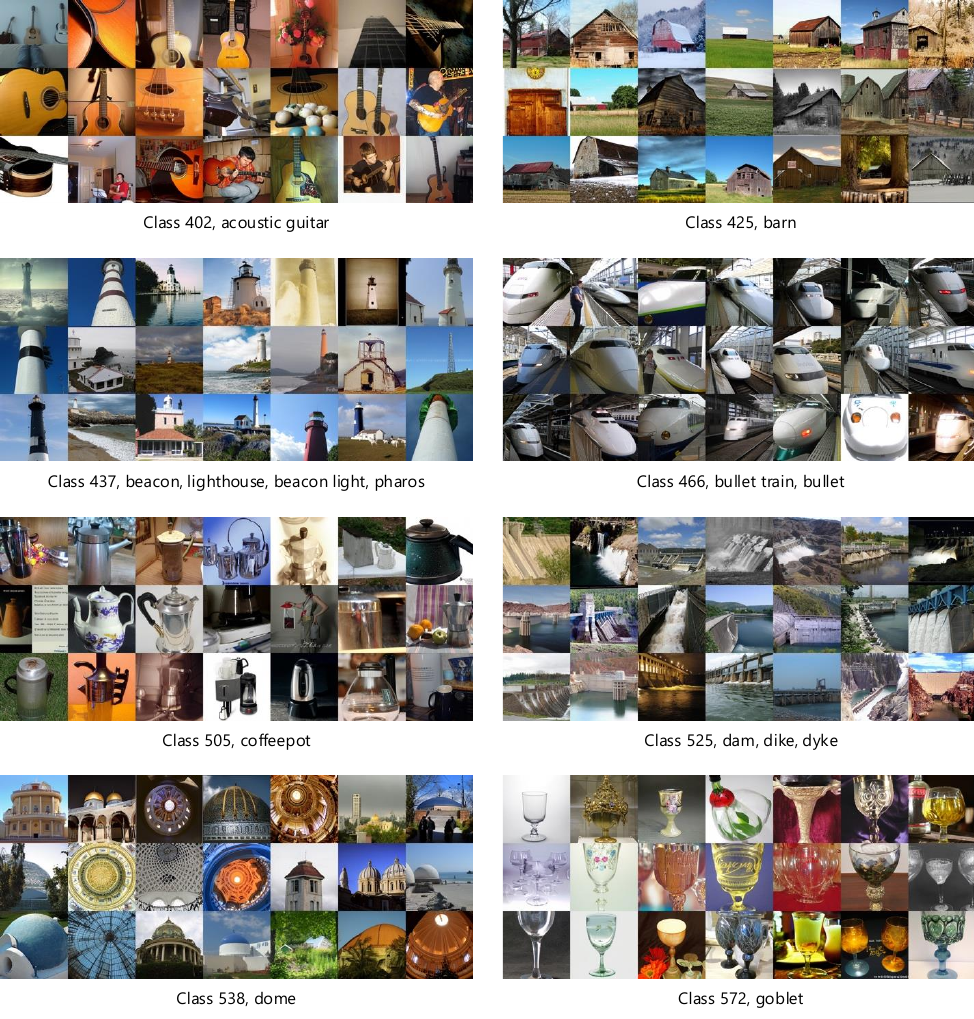}
    \caption{\textbf{Uncurated samples from EiT (480 epochs) with guidance on ImageNet $256\times256$.}}
    \label{fig:vis_dm3}
\end{figure}

\begin{figure}
    \centering
    \includegraphics[width=\linewidth]{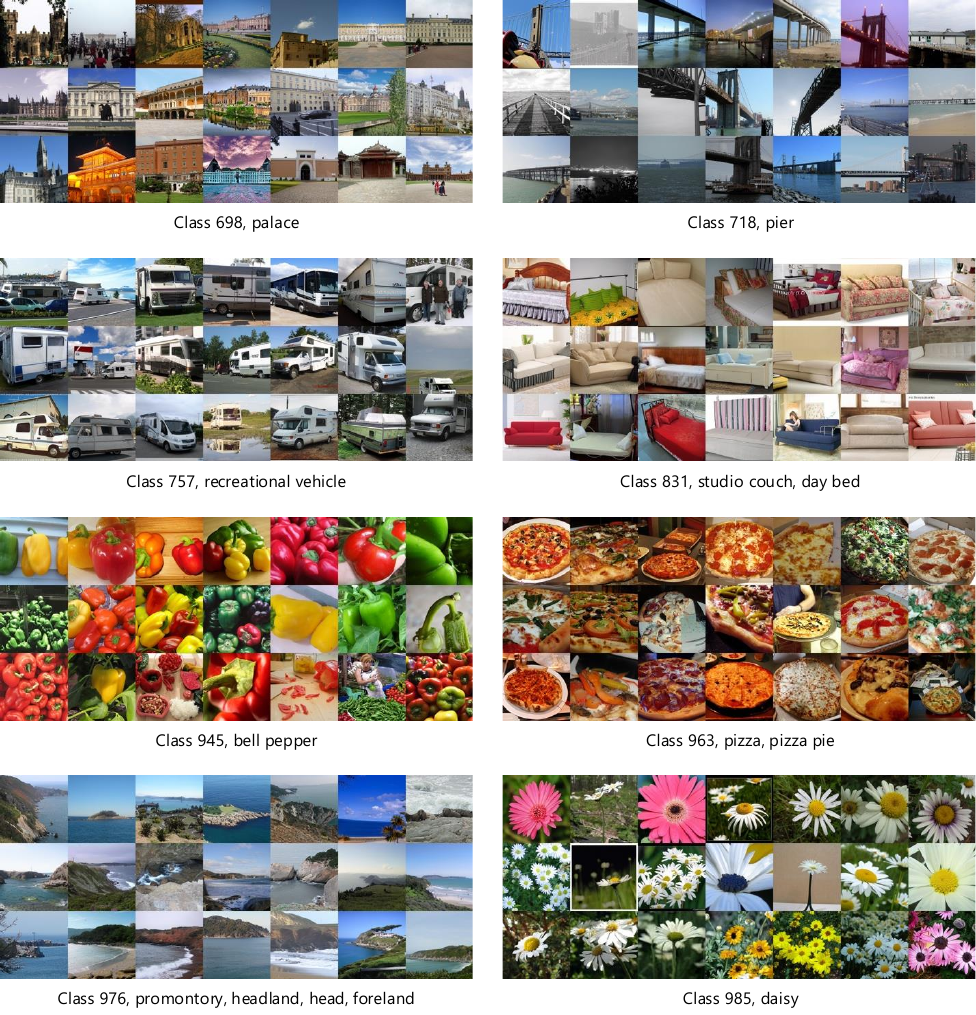}
    \caption{\textbf{Uncurated samples from EiT (480 epochs) with guidance on ImageNet $256\times256$.}}
    \label{fig:vis_dm4}
\end{figure}

\begin{figure}
    \centering
    \includegraphics[width=\linewidth]{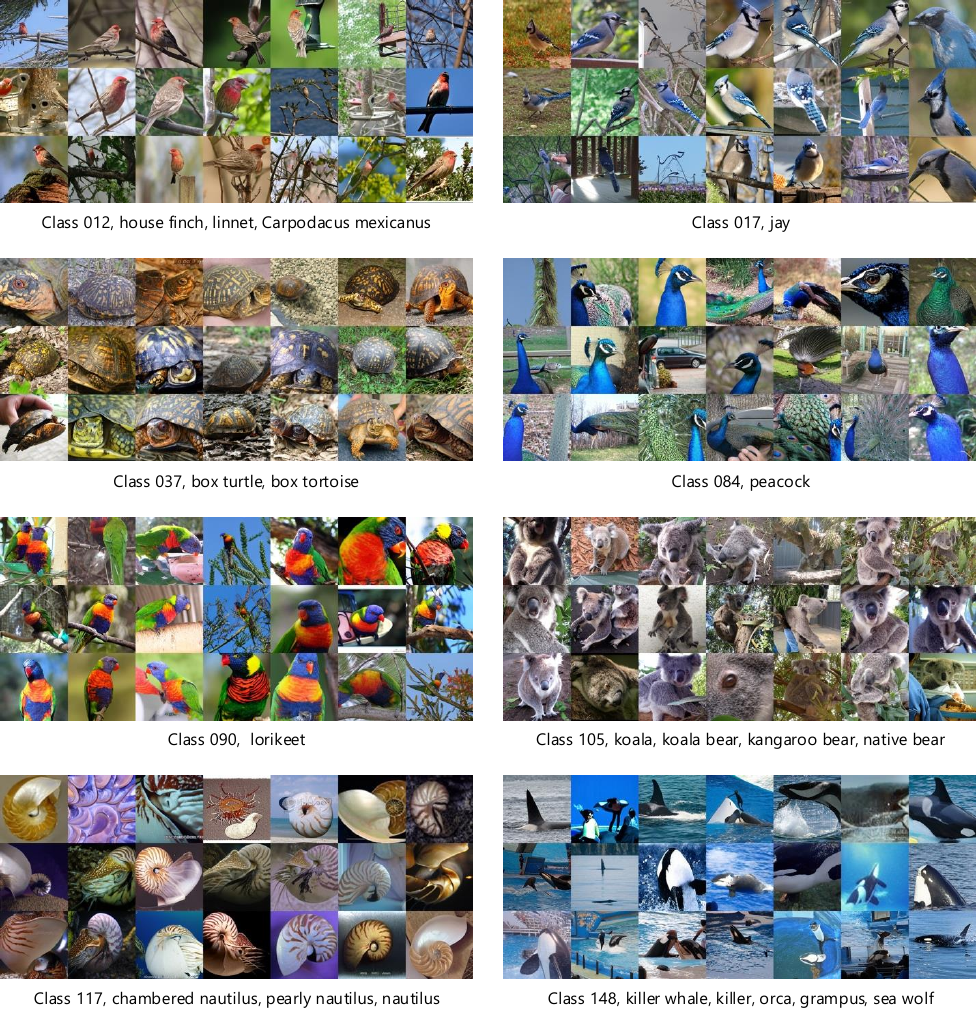}
    \caption{\textbf{Uncurated samples from EAR (640 epochs) with guidance on ImageNet $256\times256$.}}
    \label{fig:vis_ar1}
\end{figure}

\begin{figure}
    \centering
    \includegraphics[width=\linewidth]{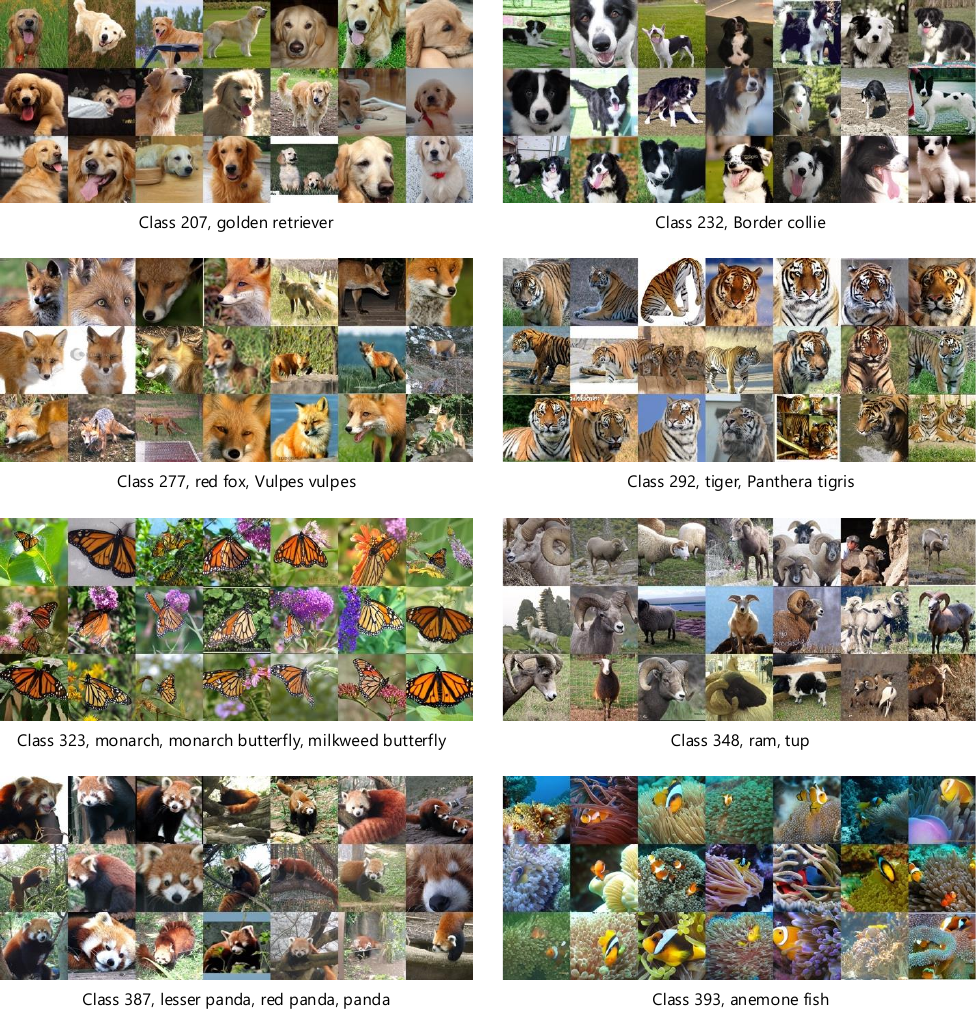}
    \caption{\textbf{Uncurated samples from EAR (640 epochs) with guidance on ImageNet $256\times256$.}}
    \label{fig:vis_ar2}
\end{figure}

\begin{figure}
    \centering
    \includegraphics[width=\linewidth]{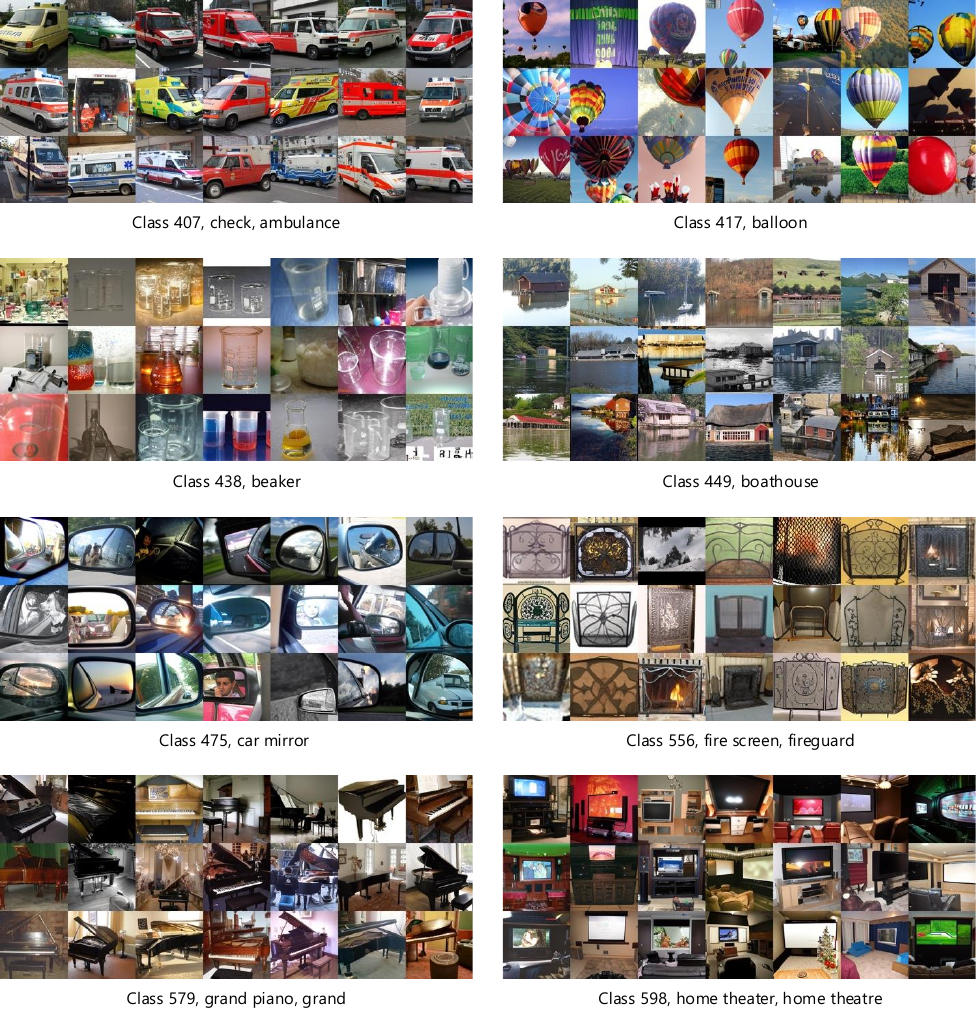}
    \caption{\textbf{Uncurated samples from EAR (640 epochs) with guidance on ImageNet $256\times256$.}}
    \label{fig:vis_ar3}
\end{figure}

\begin{figure}
    \centering
    \includegraphics[width=\linewidth]{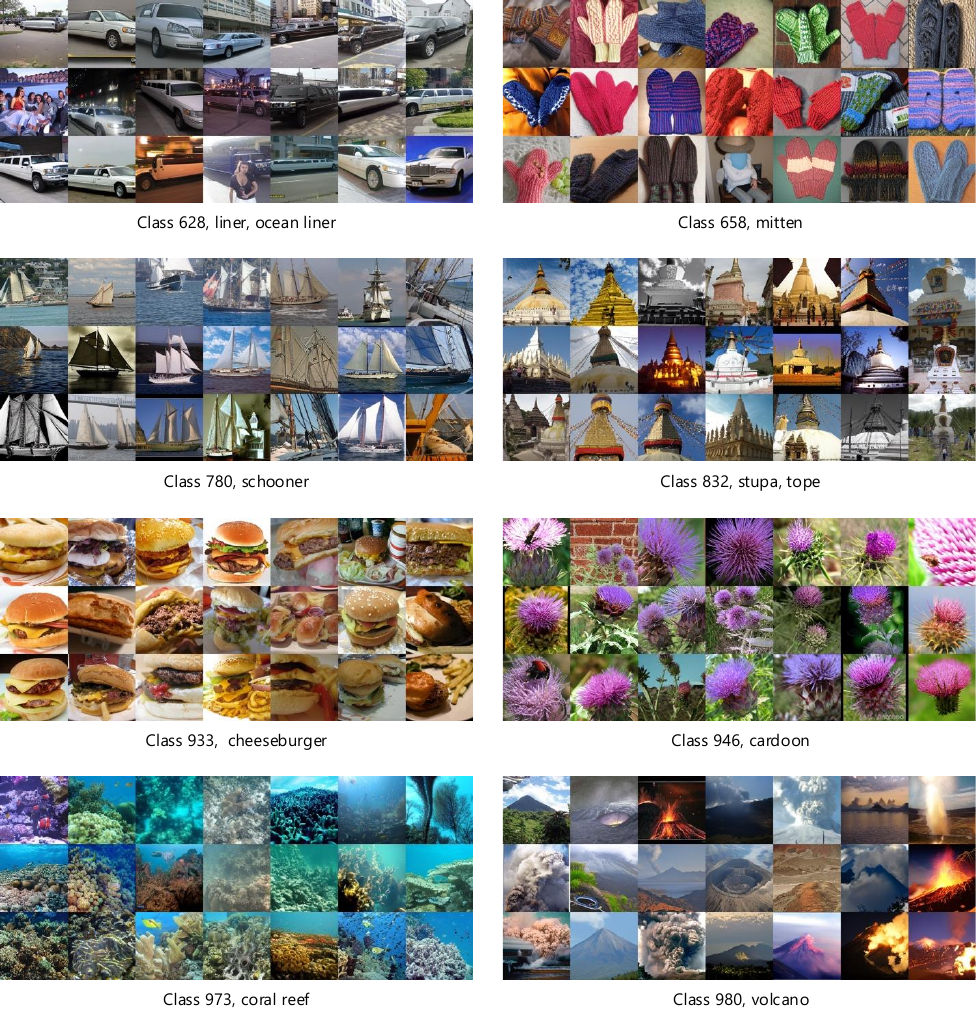}
    \caption{\textbf{Uncurated samples from EAR (640 epochs) with guidance on ImageNet $256\times256$.}}
    \label{fig:vis_ar4}
\end{figure}

\cref{fig:vis_dm1,fig:vis_dm2,fig:vis_dm3,fig:vis_dm4} present additional uncurated examples on ImageNet $256\times256$ with EiT. 
\cref{fig:vis_ar1,fig:vis_ar2,fig:vis_ar3,fig:vis_ar4} present additional uncurated examples on ImageNet $256\times256$ with EAR.

\clearpage

\bibliographystyle{plainnat}
\bibliography{main}

\end{document}